\documentclass[journal]{IEEEtran}  

\IEEEoverridecommandlockouts                              

\usepackage[utf8]{inputenc}
\usepackage[T1]{fontenc}
\usepackage{graphicx}
\usepackage{comment}
\usepackage{xcolor}
\usepackage{amsmath} 
    \usepackage{algorithm}
    \usepackage{algpseudocode}
\title{\LARGE \bf

OSCAR: An Ovipositor-Inspired Self-Propelling \\Capsule Robot for Colonoscopy }

\author{}

\author{Mostafa A. Atalla$^{1,2,\ast}$,
Anand S. Sekar$^{1}$,
Remi van Starkenburg$^{1}$,
David J. Jager$^{1}$,\\
Aim\'ee Sakes$^{1}$,
Micha\"el Wiertlewski$^{2}$,
and Paul Breedveld$^{1}$%
\thanks{$^{1}$M. A. Atalla, A. S. Sekar, R. van Starkenburg, D. J. Jager, A. Sakes, and P. Breedveld are with the Department of BioMechanical Engineering, Faculty of Mechanical Engineering, Delft University of Technology, 2628 CD Delft, The Netherlands.}%
\thanks{$^{2}$M. A. Atalla and M. Wiertlewski are with the Department of Cognitive Robotics, Faculty of Mechanical Engineering, Delft University of Technology, 2628 CD Delft, The Netherlands.}%
\thanks{$^{\ast}$Corresponding author. Email: {\tt\small m.a.a.atalla@tudelft.nl}.}%
}

\begin{document}

\maketitle
\thispagestyle{empty}
\pagestyle{empty}

\begin{abstract}
Self-propelling robotic capsules eliminate shaft looping of conventional colonoscopy, reducing patient discomfort. However, reliably moving within the slippery, viscoelastic environment of the colon remains a significant challenge. We present OSCAR, an ovipositor-inspired self-propelling capsule robot that translates the transport strategy of parasitic wasps into a propulsion mechanism for colonoscopy. OSCAR mechanically encodes the ovipositor-inspired motion pattern through a spring-loaded cam system that drives twelve circumferential sliders in a coordinated, phase-shifted sequence. By tuning the motion profile to maximize the retract phase relative to the advance phase, the capsule creates a controlled friction anisotropy at the interface that generates net forward thrust. We developed an analytical model incorporating a Kelvin-Voigt formulation to capture the viscoelastic stick--slip interactions between the sliders and the tissue, linking the asymmetry between advance and retract phase durations to mean thrust, and slider-reversal synchronization to thrust stability. Comprehensive force characterization experiments in \textit{ex-vivo} porcine colon revealed a mean steady-state traction force of $\approx 0.85~\mathrm{N}$, which closely matches the model predictions and quantifies the mechanical and viscoelastic losses. Furthermore, experiments confirmed that thrust generation is speed-independent and scales linearly with the phase asymmetry, in agreement with theoretical predictions, underscoring the capsule's predictable performance and scalability. In locomotion validation experiments, OSCAR demonstrated robust performance, achieving an average speed of $3.08~\mathrm{mm/s}$, a velocity sufficient to match the cecal intubation times of conventional colonoscopy. By coupling phase-encoded friction anisotropy with a predictive model, OSCAR delivers controllable thrust generation at low normal loads, enabling safer and more robust self-propelling locomotion for robotic capsule colonoscopy.
\end{abstract}

\section{Introduction}
Colorectal cancer (CRC) remains a major global health burden, ranking as the third most diagnosed malignancy and the second leading cause of cancer-related death worldwide, with recent estimates reporting 1.9 million new cases and nearly 1 million deaths each year~\cite{Sung2021, Siegel2024}. The most effective way to reduce this mortality is timely screening with colonoscopy. As the only modality that enables detection and removal of precancerous adenomas in a single session, colonoscopy reduces CRC incidence by up to 69\% and mortality by 68\%~\cite{Zauber2012, Doubeni2013}. Supported by this proven clinical value, more than 19 million procedures are performed annually in the United States and approximately 50 million procedures are performed each year worldwide.~\cite{Peery2022,OlympusIR2021}. Yet, despite its clinical significance, conventional colonoscopy faces fundamental mechanical limitations that increase procedural difficulty for clinicians and compromise clinical outcomes.

\begin{figure}
    \centering
    \includegraphics{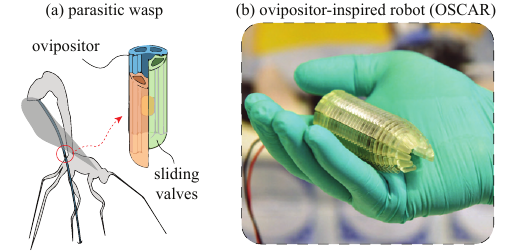}
    \caption{\textbf{Ovipositor-Inspired Capsule Robot for Colonoscopy (OSCAR).}
(a) Parasitic wasps use reciprocating sliding valves that alternately anchor and release to transport eggs.
(b) OSCAR translates this principle into a compact capsule with reciprocating sliders that generate self-propelling motion.}
    \label{fig:intro}
\end{figure}

Conventional colonoscopy relies on the manual advancement of a flexible endoscope through a tortuous and highly compliant anatomy~\cite{CottonWilliams}. Variability in colon anatomy across the population, ranging from differences in length and diameter to mobility, creates unpredictable boundary conditions that complicate instrument navigation and results in a steep learning curve for novice clinicians~\cite{saunders1996,Sedlack2011,patel2014}. Pushing the colonoscope from the proximal end frequently causes its shaft to loop, especially in mobile segments such as the sigmoid, due to the dynamic interaction between the flexible shaft and the highly deformable and stretchable nature of the colon~\cite{Shah2002, sato2006}. This looping is the main reason behind patient pain and discomfort \cite{JAhmed2024} as it subjects the colon walls to excessive forces, resulting in patient pain and forcing operators into successive motion cycles of pulling, twisting, and pushing the colonoscope forward~\cite{Shah2002} to resolve the loop and to have a path forward again. Furthermore, the control handle of conventional colonoscopes is difficult to use. Its design requires the user to carry the scope while operating two concentric knobs simultaneously to control the scope tip configuration, which is non-ergonomic and demands the operator to mentally map the rotations of two knobs to 3D tip movements while exerting insertion force, increasing cognitive load~\cite{sewell2016} and contributing to a high prevalence of musculoskeletal injuries among endoscopists~\cite{Hansel2009MSK, Lipowska2021}. In challenging cases, these mechanical and ergonomic limitations increase the risk of failed cecal intubation, necessitating deeper sedation or alternative imaging modalities~\cite{awadie2018}. These limitations underscore the need for new technologies that reduce pain, improve patient comfort, improve controls, and reduce operator stress.

Robotic approaches aim to overcome the mechanical limitations of conventional colonoscopy by relocating actuation from the proximal handle to the distal tip. Instead of pushing the instrument and inducing loop formation, robotic systems seek to pull or guide the device through the lumen, thereby reducing both forces and stretch of the colon wall. These concepts have been explored through two broad strategies: external and internal actuation. External actuation uses electromagnetic coils or permanent magnets integrated into a robot arm to manipulate and steer a capsule or flexible endoscope equipped with an internal magnetic element~\cite{Carpi2011,Valdastri2012,Chen2014,Pittiglio2019, Martin2020}. Although this approach successfully eliminates looping and enables nearly painless navigation, it demands large, costly infrastructure and suffers from force limitations in patients with a high body mass index~\cite{Martin2020, Ciuti2020}. These constraints have motivated a shift toward internal actuation approaches capable of self-propelling through on board mechanisms, with the goal of achieving the safety benefits of distal actuation without the logistical requirements of magnetic platforms.

The actuation architecture of self-propelling endoscopes follow either a wireless or a tethered design. Wireless capsule endoscopy offers a non-invasive diagnostic experience, but remains fundamentally limited by its passive propulsion, restricted onboard energy (battery life), and lack of biopsy/therapeutic capabilities~\cite{Slawinski2015, Ciuti2020, Manolescu2024_EDHT}. Consequently, achieving the functional capabilities of a conventional colonoscope still requires a physical connection. This need has driven the development of self-propelling tethered robotic capsules~\cite{Manfredi2021}, which incorporate mechanisms onboard to generate propulsion while pulling a lightweight, flexible tether supplying power, fluids, and high-quality data transmission. By combining the mobility of a capsule with the functional robustness of a tether, such devices represent a promising pathway toward painless yet fully therapeutic colonoscopy.

Several self-propelling architectures have been proposed in the literature to replace the push-driven colonoscope, including inchworm and crawler robots, track- and legged-based devices, and self-growing "vine" robots. Inchworm-type systems advance by sequentially anchoring and releasing their moving segments, typically via vacuum/suction or inflatable balloons~\cite{Manfredi2019,Manfredi2021,Kim2021Easycolon,Bernth2024, Giri2025}, and have demonstrated acceptable diagnostic performance in human trials~\cite{Cosentino2009,Tumino2010,Tumino2017}. However, their inherently discontinuous clamp–extend gait leads to a substantially long cecal intubation time of ($51 \pm 22.5~\mathrm{min}$) on average~\cite{Tumino2017}, compared to ($5-15~\mathrm{min}$) in routine practice and training benchmarks of conventional colonoscopy~\cite{Sedlack2011, kim2021cecal}, limiting their adoption beyond niche indications.

Alternatively, friction-driven robots that employ legs, paddles, or tracks can achieve higher speeds and more stable locomotion~\cite{Quirini2008,Kim2010,Valdastri2009, Consumi2023,Du2025Track}. However, in the colon, their performance is fundamentally limited by the low effective friction coefficient of the mucosa: generating sufficient traction often requires elevated normal forces, which increases the risk of mucosal trauma. In addition, tether drag, particularly for pneumatically-driven capsules, and miniaturization required for capsule designs remain major practical challenges \cite{Ortega2021B}. More recent soft self-growing colonoscope prototypes address the friction problem by growing from the tip~\cite{EndoVine,Shi2025,Kim2025SoftGrowingEndoscope, Suulker_2026}, and thus minimizing forces on the wall compared to standard colonoscopy~\cite{Ahmed2025}. However, these systems still face unresolved challenges in retraction, reliable steering, integration of tools, and dependence on bulky pneumatic infrastructure, keeping them in the prototype stage.

As a result, these limitations suggest the need for a self-propelling solution that simultaneously offers compactness, high-speed, yet robust propulsion at low normal loads, robust towing of a functional tether, and a safety profile suitable for routine screening.

Addressing this unmet need, we present OSCAR: an ovipositor-inspired capsule robot for colonoscopy (Fig.\ref{fig:intro}), a self-propelling capsule robot that takes inspiration from the way parasitic wasps drive their slender ovipositor needles through dense substrates, translating this biological strategy into a locomotion design concept suitable for the colon. This work makes four main contributions: First, we introduce a self-propelling locomotion principle capable of generating a continuous, tissue-safe thrust without relying on large normal forces. Second, we present a novel design that realizes this self-propelling principle, which we implement in a physical prototype. Third, we develop an analytical modeling framework to mathematically describe the interaction between the capsule and the tissue wall, which we use to understand OSCAR’s thrust generation mechanism, and to uncover how the design parameters, operating conditions and the capsule-tissue interaction properties (viscoelasticity and friction) affect the thrust, and to explore the design space beyond the presented prototype. Finally, we validate the locomotion principle and the mathematical model through systematic force-characterization and locomotion experiments in \textit{ex-vivo} porcine colon.

In the remainder of this paper, Section~\ref{sec:bioinspiration} examines the biological inspiration in detail, outlining the locomotion principle of the wasp ovipositor. Section~\ref{sec:design} presents the conceptual design and implementation of OSCAR, from the bio-inspired mechanism design to the physical prototype. In Section~\ref{sec:modeling}, we develop a mathematical model that captures OSCAR’s kinematics and its interaction with the colonic wall, which we then use in Section~\ref{sec:simulation} to explore the design space through simulation. Section~\ref{sec:experiment} describes the force characterization and locomotion validation experiments, and Section~\ref{sec:discuss} presents and discusses their results, implications and limitations. Finally, Section~\ref{sec:conclusion} summarizes the main findings and directions for future work.

\begin{figure*}
    \centering
    \includegraphics[width=\linewidth]{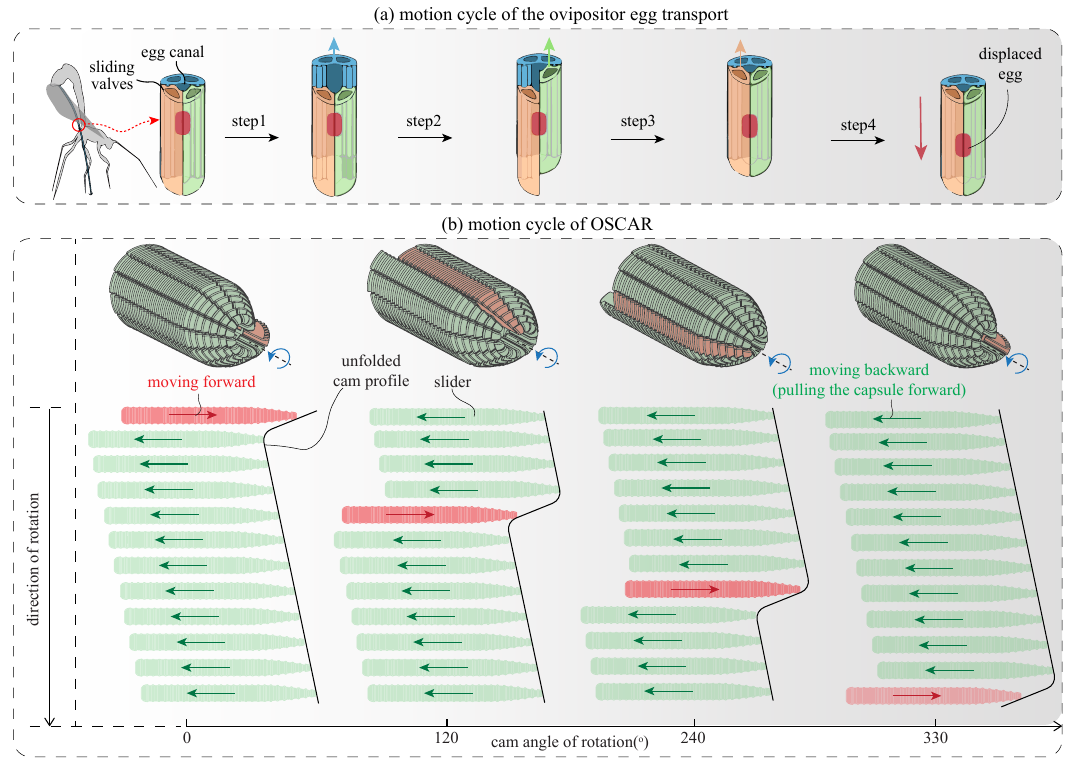}
    \caption{\textbf{Bio-inspiration and conceptual locomotion principle of OSCAR.} (a) Schematic representation of a parasitic wasp transporting eggs through coordinated stick–slip interactions generated by three sliding valves in their ovipositor. During each cycle, one valve retracts while the other two remain stationary, holding the egg in place because of the higher friction. When all valves reach their retract position, they advance together, producing a net forward displacement of the egg. Repeating this sequence enables controlled transport of the egg through the lumen till its deposition. (b) OSCAR translates and extends this principle into a capsule architecture using circumferential sliders driven by a rotating cam profile. Unlike the biological mechanism, where one valve moves at a time while the rest remain stationary, OSCAR drives a slider forward while driving the remaining sliders in the backward during each stroke. This continuous, phase-shifted reciprocation can yield a steady and continuous steadier average forward thrust suitable for self-propelling locomotion in colonoscopy.}
    \label{fig:working_principle}
\end{figure*}

\section{Bio-inspiration from Wasp Ovipositor}\label{sec:bioinspiration}
Biological systems have evolved efficient strategies for locomotion through confined substrates, exemplified by the ovipositor of parasitic wasps. The ovipositor is an exceptionally slender, flexible organ that allows the wasp to penetrate substrates and deliver eggs deep inside host tissues (Fig.~\ref{fig:working_principle}(a)). Its functionality relies on a set of three longitudinal valves, one dorsal and two ventral, arranged in a tongue-and-groove configuration that mechanically locks them together radially while allowing independent sliding along their length~\cite{deKater2021FlexibleWaspTransport, AUSTIN, Cerkvenik2017OvipositorSteering, vanMeer2020OvipositorActuation, Ahmed2013OvipositorSenseOrgans}.

A key feature of this biological system is its ability to achieve controlled transport within a narrow tubular conduit without relying on large normal forces. Current theories suggest that the wasp induces frictional anisotropy between the valves and the egg to create net transport motion by coordinating the motion sequence of its valves~\cite{AUSTIN, Ahmed2013OvipositorSenseOrgans}. During each cycle, one valve retracts while the other two remain stationary to maintain higher frictional grip on the egg, effectively locking it in place. The motion then alternates sequentially between sliders until all sliders are in retraction position, after which all three slide forward together, advancing the egg by a stroke length. This coordinated stick–slip sequence produces reliable forward transport using only internal reciprocating motions. The wasp ovipositor, therefore, exemplifies a low-force strategy for propulsion in confined spaces, well suited for self-propelling endoscopic systems. 

The transport concept of wasp ovipositors has inspired several medical and robotic innovations, including self-propelling needles that advance through soft tissue using interlocking sliding segments~\cite{Leibinger2016MinimallyDisruptive,Bloemberg2024}, friction-based tissue-transport mechanisms that move samples through narrow channels~\cite{Sakes2020TissueTransport}, and soft robots for pipeline inspection~\cite{Atalla2024MechanicallyInflatable}. Collectively, these examples highlight the broad applicability of ovipositor-inspired locomotion and its potential to address propulsion challenges in colonoscopy.

\section{Design \& Implementation of OSCAR}\label{sec:design}

Inspired by the ovipositor’s strategy of generating locomotion through coordinated sliding of its interlocked valves, OSCAR translates this biological principle into a capsule-scale propulsion concept suitable for the colon. The core idea is to distribute multiple longitudinal sliders evenly around the capsule body and drive their reciprocating motion using a rotating cam, as shown in Fig.~\ref{fig:working_principle}(b). As the cam turns, each slider follows a prescribed phase-shifted trajectory, which sequences its forward and backward strokes relative to the others. This cam-driven coordinated motion enables a single drive input for all sliders and allows for steady thrust generation by simultaneously moving all sliders, in contrast to the discrete stepwise gait of the ovipositor.

The net forward thrust requires an asymmetry in the wall interaction such that, at any instant, more sliders retract than advance ($n_{ret}>n_{adv}$), analogous to the ovipositor where one valve advances while the others hold position. This phase asymmetry creates an anisotropic frictional condition at the capsule–wall interface: the retracting sliders collectively generate a larger friction force that pushes the capsule forward, than the opposing friction produced by the advancing sliders, such that $\sum F_{ret} > \sum F_{adv}$, with $\sum F_{ret} = n_{ret} F_f$ and $\sum F_{adv} = n_{adv} F_f$, where $F_f$ is the friction force of a single slider. The resulting friction imbalance yields a nonzero net forward thrust, enabling self-propulsion in the same way that coordinated valve sliding transports eggs along the ovipositor. As such, achieving this force imbalance requires a minimum of three sliders, such as in the case of the wasp ovipositor. 

Beyond nature’s minimalist design, increasing the number of sliders is desirable, as it can further bias the cycle toward more retracting than advancing sliders and thereby strengthen the friction anisotropy and its resulting force imbalance, potentially yielding higher net thrust. Consequently, maximizing the number of sliders of the capsule becomes an important design consideration.

\begin{figure*}
    \centering
    \includegraphics[width=0.9\linewidth]{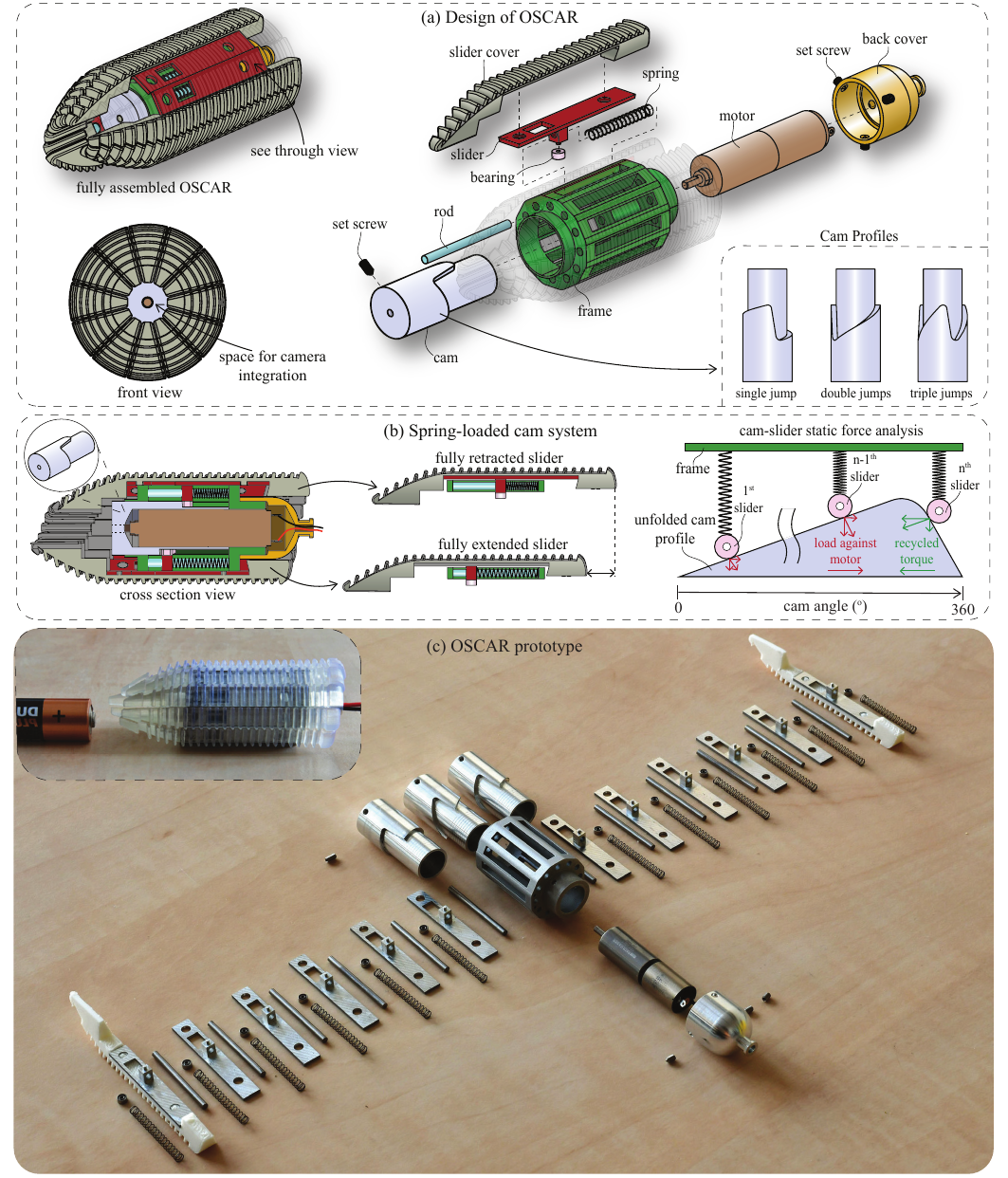}
    \caption{\textbf{Design and prototype of the OSCAR capsule robot.}
    (a) The design of OSCAR consists of 12 spring-loaded sliders distributed evenly around the circumference of the capsule body and driven by a central cam mechanism to realize the bio-inspired motion cycle described in the working principle. Each slider is guided by a cylindrical rod to ensure purely axial motion and is covered by a grooved slider cap to enhance frictional contact with the mucosa. The cam is mounted concentrically on a motor and enclosed by a back cover. Different cam profiles with varying numbers of motion reversals (jumps) were designed and fabricated to study the influence of motion pattern on thrust. 
    (b) The spring-loaded cam system enables the use of steep cam angles while maintaining continuous contact between the sliders and the cam profile. This configuration allows maximum number of sliders to remain engaged in retraction, maximizing the potential net thrust. During the advance (fall) phase, the spring-loaded force of the advancing slider acting on the cam profile recycles the stored spring energy by generating a torque that assists the motor, thereby reducing the required actuation load. 
    (c) Assembled and disassembled prototypes of OSCAR, showing all key components and the compact integration of the actuation and transmission mechanisms.
    }
    \label{fig:proto}
\end{figure*}

Overall, this conceptual design establishes a sliding system architecture that captures the essence of ovipositor-inspired locomotion while adapting to the colonoscopy application quest for steady and stable propulsion.

\subsection{Design Requirements}
The design of OSCAR is guided by a set of anatomical and mechanical requirements that the capsule must satisfy to function reliably inside the human colon. The colon diameter varies substantially along its length, ranging from approximately $30~\mathrm{mm}$ in the sigmoid colon to over $80~\mathrm{mm}$ in the cecum~\cite{Alazmani2016}. To ensure safe passage through bends and local constrictions, the capsule diameter must therefore remain below the minimum diameter of the sigmoid colon (< $30~\mathrm{mm}$).

In addition to the size requirement, the system must meet practical procedure-time targets. 
To achieve a cecal intubation time comparable to conventional colonoscopy ($5-15~\mathrm{min}$)~\cite{Sedlack2011, kim2021cecal}, 
the capsule should traverse a colon of $\sim 1.5~\mathrm{m}$ median length (and potentially $>1.8~\mathrm{m}$)~\cite{Alqarni2024ColonLength}, 
which implies an average forward speed on the order of $\sim 2$--$6~\mathrm{mm/s}$.

The propulsion efficiency is further constrained by the tribological properties of the colon. The mucosa is coated with a lubricating mucus layer, whose coefficient of friction with the capsule can range from very low $\mu \approx 0.03$ to relatively high $\mu \approx 0.5$, depending on the material and design of the capsule sliders, the lubricity of the colon wall, and the normal load between the capsule and the colon wall~\cite{Accoto2001WTC, Norton2020,  Jinyang2024}. To achieve locomotion, the capsule must generate net thrust force that overcomes the static friction with the wall, while also carrying its own load and counteracting any additional resistance sources such as tether drag. For similar capsules, the net thrust force reported is typically around $1.0~\mathrm{N}$~\cite{Jinyang2024}.

These performance targets must remain within strict safety boundaries. Local contact shear forces should stay within safe limits to avoid superficial mucosal trauma and, critically, prevent damage beyond the mucosal layer \cite{Norton2020}. The propulsion mechanism must therefore balance traction generation with tissue protection, producing consistent forward force while minimizing local pressure concentrations. 

Together, these geometric, kinematic, and tribological requirements define the design space for OSCAR.

\subsection{Detailed Design of OSCAR}
Building on our conceptual design and taking into account the design requirements, we strived for a capsule that complies with the size constraint (\(<30~\mathrm{mm}\) diameter) while still maximizing the number of sliders to increase the maximum attainable thrust, as implied by our design concept. We therefore selected a capsule diameter of \(28~\mathrm{mm}\) to leave a safety margin below the size limit. Within this diameter, we were able to fit a maximum of twelve sliders while keeping the design realistic to fabricate and assemble; beyond this number, the feature sizes and tolerances become hard to manufacture. The sliders are distributed evenly around the circumference of the frame and are constrained to translate axially along dedicated grooves machined into the main frame, as illustrated in Fig.~\ref{fig:proto}(a).

Each slider carries a cover to interface with the colon wall. The cover features a slotted surface design to enhance frictional grip against the intestinal wall by trapping mucus and pushing any fluid at the interface into the slots, and leveraging local deformation of the colon wall between the slots to achieve better grip~\cite{Norton2020}. The slot ridges are symmetric (i.e., not inclined in either direction) to ensure that sliding friction is symmetric. The cover is rounded (oval) at both ends to ensure smooth interaction at the capsule boundaries.

To implement the alternating retract-advance pattern inspired by the ovipositor mechanism, the cam profile actuating the sliders was designed to maximize the instantaneous imbalance between retracting and advancing sliders, thereby strengthening the motion asymmetry and the potential net thrust. In addition, we strived to provide sufficient stroke length to overcome the elastic wall reaction and break the static friction, which is necessary to propel forward. We defined the slider stroke length to be ($10~\mathrm{mm}$).

Producing a cam profile that achieves this stroke length while maximizing the ratio of retracting to advancing sliders requires steep motion reversal angles. For example, enabling a maximum of eleven sliders to retract while only one advances implies that the advance phase occupies \(1/12\) of the cam cycle, requiring reversal angles \(>80^\circ\) to push forward the slider in the advance phase. Under these conditions, traditional grooved cam–follower designs are not feasible: such steep reversals would require very high actuation torque to force the followers through the corners, or otherwise the sliders would be mechanically-locked and the motion would stall. To overcome this challenge, we designed a single-sided cam profile (rather than a closed groove), where the sliders are preloaded against the cam surface with springs to maintain contact as the cam rotates. This spring-loaded cam--slider design prevents mechanical locking and simplifies the torque requirements.

A positive side effect of this design is that the slider(s) in the advance phase can recycle the spring energy accumulated during retraction by pushing against the cam wall, generating an assisting torque that supports the motor and partially relieves it from the load, as illustrated in Fig.~\ref{fig:proto}(b). In this configuration, the steep reversal angle becomes beneficial because they increase the force component that contributes to this assisting torque. For experimental validation, three cam profiles were developed with one, two, and three motion reversals (``jumps'') per revolution, as shown in Fig.\ref{fig:proto}(a), enabling systematic variation of the advance and retract timing. Each cam was machined with the motion profile around its circumference while being hollow to house the driving motor, reducing the axial length of OSCAR and making the device more compact, as shown in Fig. 3(b). We selected a pre-loading spring of \(5~\mathrm{N/mm}\) stiffness to ensure reliable tracking of the cam profile under varying tissue stiffness and surface lubricity conditions.

The final OSCAR design measures \(28~\mathrm{mm}\) in diameter and \(60~\mathrm{mm}\) in length, with a slider stroke length of \(10~\mathrm{mm}\) and a total assembled mass of approximately \(50~\mathrm{g}\), as indicated in Table~\ref{tab:design_parameters}. The overall architecture is modular, allowing rapid cam exchange and straightforward assembly while preserving space at the anterior end of the capsule for potential imaging integration.

\subsection{Prototype Development}
The prototype of OSCAR was developed to validate its working principle and assess its locomotion performance. Structural components were fabricated using milling and electric discharge machining (EDM) from aluminium 7075–T5, which we selected due to its high strength-to-weight ratio and dimensional stability under load. To accommodate the intricate features of slider covers, these covers were additively manufactured using stereolithography (SLA) and subsequently bonded to the machined sliders. Standard off-the-shelf components—including compression springs, guide rods, and miniature bearings—were integrated to streamline assembly and facilitate future iterations. The prototype was driven by a 12-V DC motor with an integrated gearbox to deliver the torque required to operate the capsule (1016M012S, FAULHABER, Schönaich, Germany). Figure~\ref{fig:proto}(c) shows both the assembled and disassembled states of the prototype, highlighting the compact arrangement and the mechanical interfaces between the frame, cam, motor, sliders, and springs.

\begin{table}[b!]
    \centering
    \caption{Design parameters of the OSCAR prototype.}
    \label{tab:design_parameters}
    \begin{tabular}{lll}
        \hline\hline
        \textbf{Design parameter} & \textbf{description} & \textbf{Value} \\ 
        \hline
        $D_{capsule}$        & Capsule diameter            & $28~\mathrm{mm}$ \\
        $L_{capsule}$      & Capsule length              & $60~\mathrm{mm}$ \\
        $h_{stroke}$        & Slider stroke               & $10~\mathrm{mm}$ \\
        $m$        & Mass                        & $50~\mathrm{g}$ \\
        \hline\hline
    \end{tabular}
\end{table}

\section{Modeling of OSCAR}\label{sec:modeling}
In order to understand the thrust generation mechanism of OSCAR and its contributing factors, we built an analytical model that captures the kinematics of a single slider and its force interaction with the wall, which we then extended to the full capsule thrust by superimposing the phase‐shifted contributions of the $n$ sliders. 

\subsection{Single Slider Kinematics}
\begin{figure}[t!]
    \centering
    \includegraphics[width=\linewidth]{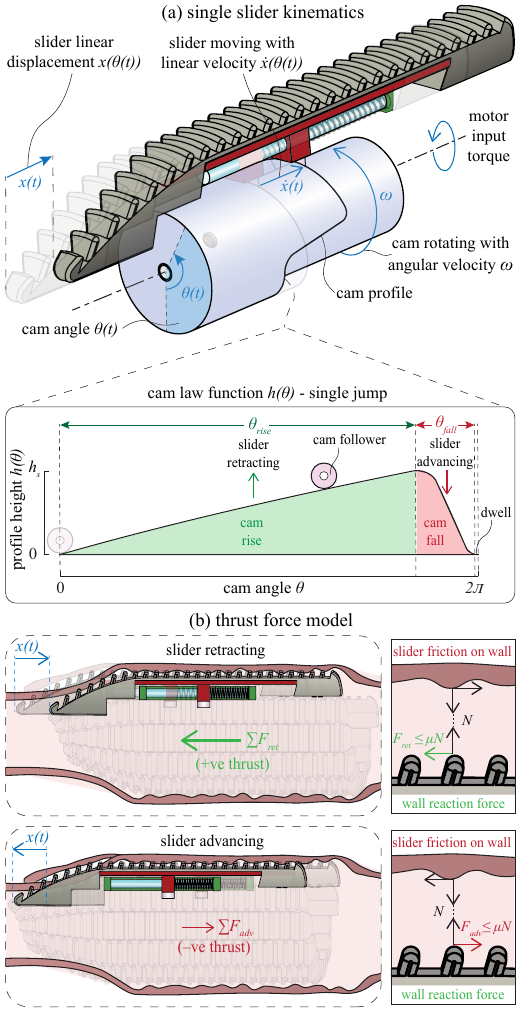}
    \caption{\textbf{Single--slider kinematics and thrust force generation.}
(a) Kinematic description of OSCAR’s slider actuation. A cam rotating with angular velocity~$\omega$ drives each slider through a prescribed linear trajectory $x(t)$. The cam law for the single--jump profile consists of a slow rise phase (slider retraction), a steep fall phase (slider advance), and a short dwell. The interval asymmetry between rise and fall sets the duty--cycle imbalance that underpins net thrust generation. (b) Thrust force model based on frictional interaction with the colonic wall. During retraction (top),
the slider moves backward relative to the capsule, generating positive thrust as the wall traction opposes the slider motion. During advance (bottom), the slider moves forward, producing
negative thrust as the wall reaction reverses direction. Superposition of these phase--shifted contributions of all sliders yields the average net thrust force.}
\label{fig:model}
\end{figure}

Consider a single slider from the actuation system of OSCAR, spring–loaded against the cam as illustrated in Fig.~\ref{fig:model}(a). We assume that the spring preload maintains continuous contact between the slider and the cam surface, such that separation or backlash between the slider and the cam profile does not occur during the cam cycle. Under this assumption, the slider is kinematically constrained to the cam which means that its axial displacement is fully prescribed by the cam profile and does not depend on inertia or spring dynamics. This assumption holds when the preload force exceeds the resisting forces due to slider inertia during cam actuation and due to slider–wall interaction.

The cam profile is described by a piecewise lift function $h(\theta)$, which gives the imposed slider displacement as a function of the cam angle $\theta$. In all cases, the cam law is expressed as a continuous piecewise function composed of linear and circular segments that represent the rise, fall, and dwell portions of the cam (Fig.~\ref{fig:model}). This can be written in general form as follows:

\begin{equation}\label{eq:displacementfunction}
h(\theta) =
\begin{cases}
h_1(\theta), & \theta_0 \le \theta < \theta_1, \\[4pt]
h_2(\theta), & \theta_1 \le \theta < \theta_2, \\[4pt]
h_3(\theta), & \theta_2 \le \theta < \theta_3, \\[4pt]
\ldots       & \text{until } 2\pi,
\end{cases}
\end{equation}

where $h_i(\theta)$ denotes the $i$-th segment of the cam law defined on the angular interval $[\theta_{i-1}, \theta_i)$, with $0 \le \theta_0 < \theta_1 < \theta_2 < \cdots \le 2\pi$. Each $h_i$ corresponds to either a circular arc or a linear segment as defined by the cam design.

Assuming that the cam rotates with a constant rotation speed $\omega$, the cam angular displacement $\theta(t)$ can be described as follows:

\begin{equation}\label{eq:camAngle}
\theta(t)=\omega t,\qquad \theta\in[0,2\pi].
\end{equation}

Since the slider is kinematically constrained to the cam, we can find the slider displacement corresponding to the cam rotation by substituting the cam angle expression Eq.~\ref{eq:camAngle} into the cam profile lift function Eq.~\ref{eq:displacementfunction}, as follows:

\begin{equation}
x_{slider}(t) = h(\theta(t))
\end{equation}

By applying the chain rule and differentiating the slider displacement $x_{slider}(t)$ in time, we can find the corresponding slider velocity:
\begin{equation}\label{eq:sliderVel}
\dot x_{slider}(t)= \frac{dh}{d\theta}\,\frac{d\theta(t)}{dt}
   = \omega\,\frac{dh}{d\theta}(\theta(t))
\end{equation}
We can see from Eqs.~(\ref{eq:camAngle})–(\ref{eq:sliderVel}) that assuming a constant cam rotation speed $\omega$ results in time-invariant kinematics, which means that the slider motion depends solely on the cam angle $\theta$ and not explicitly on time. Consequently, for every identical angular position of the cam, the slider possesses the same displacement, velocity, and acceleration states.

\subsection{Single Slider Instantaneous Force}
When the capsule interfaces with the colon wall through the sliders, the wall and the capsule press against each other with a normal load $N$, which can originate, for example, from the deflated baseline configuration of the colon or from intra–abdominal pressure acting on the wall. Each individual slider is therefore subjected to this normal force. We define forward capsule thrust as positive. When the slider retracts, meaning the slider moves backward relative to the capsule, the wall pushes forward on the capsule with a reaction force $F_{\mathrm{ret}}$. Conversely, when the slider advances, meaning the slider moves forward relative to the capsule, the wall reaction is directed backward and is denoted $F_{\mathrm{adv}}$. Both forces, $F_{ret}$ and $F_{adv}$, are limited by the available friction at the interface.

Given our slotted slider design which helps pushing mucus and interface fluids to the slots while maintaining contact with the ridges, we assume that frictional interaction between the capsule and the wall follows the Coulomb friction law described by the coefficient of friction $\mu$ between the slider and the colon wall and the normal force at the interface $N$, in line with prior capsule–tissue models \cite{Kim2006TL, Sliker2016, Manfredi2019, Barducci2020}. Accordingly, the retraction and advance forces are bound by the friction limit at the interface, as follows:

\begin{equation}\label{eq:coulomb}
\left| F_{\mathrm{adv}} \right|,\,\left| F_{\mathrm{ret}} \right| \;\leq\; \mu N,
\end{equation}

The instantaneous slider traction force ($F_{slider}$) alternates between the retraction and advance forces depending on its instantaneous velocity (i.e. motion direction, retracting or advancing) during the cycle, which can be expressed mathematically as follows:

\begin{equation}
F_{slider}(\theta)=
\begin{cases}
\leq\; +\mu N, & \dot{x}_{slider}<0 \quad \text{(retraction)},\\[4pt] \geq\; -\mu N
, & \dot{x}_{slider}>0 \quad \text{(advance)},\\[4pt]
0, & \dot{x}_{slider}=0 \quad \text{(dwell)}.
\end{cases}
\label{eq:Fs_cases}
\end{equation}

Depending on the sliding state, sticking or slipping, the instantaneous slider traction force is bounded by $|F_{slider}| \leq \mu N$. When the slider sticks, i.e. it does not move relative to the colon wall, the slider traction force remains within the bounds of the friction limit $|F_{slider}| \leq \mu_s N$, governed by the static coefficient of friction, $\mu_s$. When the slider slips, meaning that it moves relative to the colon wall, the slider traction force saturates at the friction limit $|F_{slider}| = \mu_k N$, which is governed by the kinetic coefficient of friction $\mu_k$ where ($\mu_k < \mu_s$). These two sliding conditions that define the friction coefficient can be expressed mathematically as follows:

\begin{equation}
\mu =
\begin{cases}
\mu_s, & \dot{x}_{slider}^{wall}=0 \quad \text{(stick)},\\[4pt]
\mu_k, & \left| \dot{x}_{slider}^{wall} \right| >0 \quad \text{(slip)}.\\[4pt]
\end{cases}
\label{eq:mu_cases}
\end{equation}

Where $\dot{x}_{slider}^{wall}$ is the velocity of the slider relative to the colon wall.

The idealized Coulomb friction law in Eq.\ref{eq:Fs_cases} assumes that the contact force switches instantly between positive and negative at motion reversals. However, in reality, the compliance of the colon wall and the micro-slip at the slider-wall interface lead to gradual force transitions instead of abrupt reversals. This behavior can be captured by introducing a smoothing function $f(\lambda)$ to the instantaneous slider force $F_{slider}$, where $\lambda$ is a smoothing parameter that controls how smooth the transition is. This function matters only in the fractional portion of the cycle over which the slider reverses its direction of motion. The instantaneous slider force can therefore be expressed in a generalized form as:

\begin{equation}
F_{slider}(\theta) = \mu N\, f(\lambda,\theta)
\end{equation}

where $f(\lambda)$ varies between $+1$ and $-1$ over the transition period. Assuming $\lambda \in [0,1]$, the limit $\lambda = 0$ recovers the instant force reversal, while larger values of $\lambda$ correspond to smoother force reversals. This parameter $\lambda$ can be regarded as an empirical parameter that captures the combined effect of contact relaxation, material compliance, and lubrication on the transition smoothness. 

\begin{figure*}
    \centering
    \includegraphics[width=\linewidth]{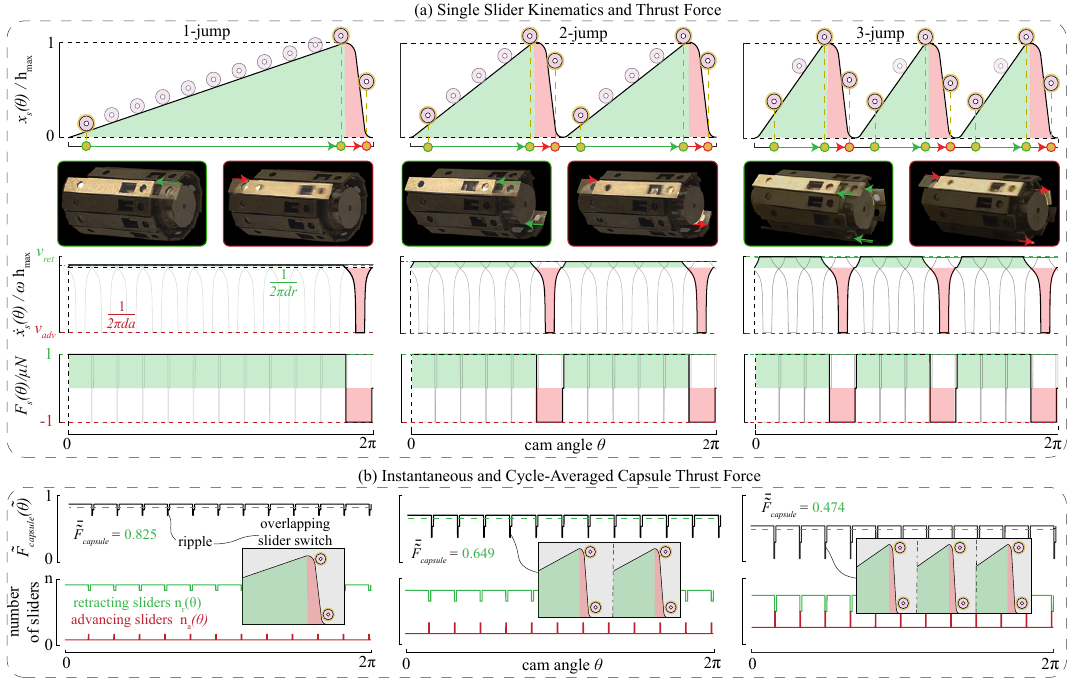}
    \caption{\textbf{Simulation of kinematics and thrust force of OSCAR.}
    (a) Single-slider displacement, velocity, and instantaneous traction force for 1-, 2-, and 3-jump cam geometries. Each cam produces distinct rise–return sequences corresponding to retracting (green) and advancing (red) phases, with normalized slider velocity given by $\dot{x}_{slider}^*/(\omega h_{\max}) = \pm 1/(2\pi d_{r,a})$.
    (b) Instantaneous and cycle-averaged capsule thrust force obtained by superposing the phase-shifted contributions of $n=12$ sliders. The upper plots show the total normalized thrust $\tilde{F}_{\mathrm{capsule}}(\theta)$ and its mean value $\bar{\tilde{F}}_{\mathrm{capsule}}$, while the lower plots show the instantaneous number of retracting $n_{ret}(\theta)$ and advancing $n_{adv}(\theta)$ sliders. Increasing the number of jumps compresses the individual stroke, increases switching frequency, and reduces the average thrust, while amplifying ripple due to increased overlapping between slider transitions. Refer to Supplementary Video.1 for a demonstration of the motion cycles of the three cam profiles.
    }
    \label{fig:simulation1}
\end{figure*}

\subsection{Single Slider Cycle-Averaged Force}
To describe the periodic force generated by a slider in a cam cycle, we define duty fractions $d_{adv}$ and $d_{ret}$, which represent the spatial fractions of the cycle spent in advance and retraction, respectively. These fractions are determined by the cam profile (including any dwells) and satisfy ($0 \leq d_{adv},d_{ret} \leq 1$) with ($d_{adv}+d_{ret} \leq 1$). The cycle–averaged thrust of a single slider can then be expressed as
\begin{equation}\label{eq:single_{adv}vg}
\bar F_{slider} \;=\; F_{\mathrm{ret}}\,d_{ret} \;-\; F_{\mathrm{adv}}\,d_{adv} \;-\; F_{\mathrm{loss}},
\end{equation}
where $F_{\mathrm{loss}}$ represents internal mechanical losses such as friction of the slider guide.

\subsection{Multi–Slider Instantaneous Force Superposition}
Assuming the capsule employs $n$ identical sliders that are uniformly phase–shifted by $2\pi/n$ around the cam profile. The instantaneous net thrust of the capsule is the superposition of all slider contributions:
\begin{equation}
\label{eq:superposition}
F_{\mathrm{capsule}}(\theta) \;=\; \sum_{i=1}^{n} F_{slider}(\theta)=\; \sum_{i=1}^{n} F_{slider}\Bigl(\theta-\tfrac{2\pi(i-1)}{n}\Bigr).
\end{equation}

Because the sliders are evenly distributed in phase, at any cam angle a certain number will be in advance and the rest in retraction. The ratio between advancing and retracting sliders is determined by the duty fractions $(d_{adv},d_{ret})$ introduced earlier. If a single slider spends a fraction $d_{adv}$ of its cycle in advance and a fraction $d_{ret}$ in retraction, then approximately the same fractions of the $n$ sliders will be in those phases at any instant, which can be expressed as follows:

\begin{equation}
n_{adv}(\theta) \;\approx\; n\,d_{adv}, 
\qquad
n_{ret}(\theta) \;\approx\; n\,d_{ret},
\end{equation}\label{eq:nr_na}
With small deviations during short overlap intervals where transitions occur. 

\subsection{Steady–State Cycled-Averaged Thrust Force}
To find the steady–state thrust of the robot, we compute the cycle–averaged force over one cam revolution by applying the averaging operator to the instantaneous capsule force expression in \eqref{eq:superposition} as follows:

\begin{equation}
 \bar F_{\mathrm{capsule}}
= \frac{1}{2\pi}\!\int_{0}^{2\pi}\!
\sum_{i=1}^{n} F_{slider}\!\Bigl(\theta-\tfrac{2\pi(i-1)}{n}\Bigr)
\,\mathrm{d}\theta.   
\end{equation}

Because the sliders are identical and uniformly phase–shifted, each term contributes the same average force, yielding:

\begin{equation}
\bar F_{\mathrm{capsule}} = n\,\bar F_{slider},
\end{equation}

where $\bar F_{slider}$ is the cycle-averaged thrust of a single slider over one cam cycle. By substituting the single slider cycle-averaged force Eq.\eqref{eq:single_{adv}vg} into the n-sliders capsule cycle-averaged thrust force expression, we find the following expression for the capsule cycle-averaged thrust force: 

\begin{equation}\label{eq:averageThrust}
\bar F_{\mathrm{capsule}} \;=\;
n\Bigl(F_{\mathrm{ret}}\,d_{ret} \;-\; F_{\mathrm{adv}}\,d_{adv} \;-\; F_{\mathrm{loss}}\Bigr)
\end{equation}

This formula shows that the cycle-averaged thrust force, that OSCAR produces, is governed by the number of sliders $n$, the geometric design of the cam profile represented by its duty fractions $d_{ret}$, $d_{adv}$, the available traction at the interface $\mu N$ represented by $F_{ret}, F_{adv}$, and the internal losses per slider $F_\mathrm{loss}$. 


\begin{figure}
    \centering
    \includegraphics[width=\linewidth]{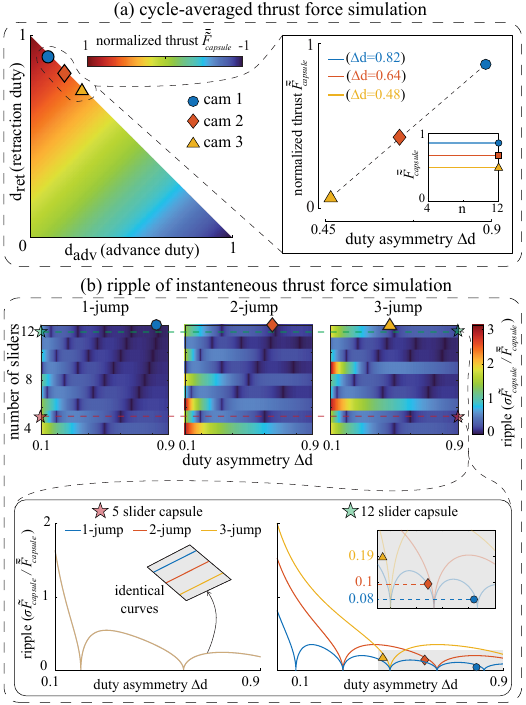}
    \caption{
\textbf{Simulation of cycle-averaged thrust and instantaneous thrust ripple of OSCAR.}
(a) \emph{Duty-fractions map} of the cycle-averaged normalized thrust $\bar{\tilde{F}}_{\mathrm{capsule}}$ over the cam duty fractions $(d_{ret},d_{adv})$. The map shows that the mean thrust is set by the imbalance between the retract and advance intervals: $d_{ret}>d_{adv}$ yields forward (positive) thrust, $d_{ret}<d_{adv}$ yields backward (negative) thrust, and $d_{ret}=d_{adv}$ yields zero mean thrust. Markers indicate the three implemented cam profiles. The plot on the right shows that the normalized cycle-averaged thrust increases linearly with the duty asymmetry $\Delta d=d_{ret}-d_{adv}$, independent of number of sliders $n$ (inset) and the number of cam motion reversals (jump count) $k$. (b) Ripple of the normalized instantaneous thrust $\tilde{F}{\mathrm{capsule}}(\theta)$, quantified as the coefficient of variation $\sigma{\tilde{F}{\mathrm{capsule}}}/\bar{\tilde{F}}{\mathrm{capsule}}$, for different slider counts $n$ and jump counts $k$. Ripple arises from overlapping slider phase switches that temporarily change the numbers of sliders in retraction and advance. The heat maps and representative curves show that $\Delta d$ shifts the phase boundaries within each of the $k$ segments relative to the slider spacing $2\pi/n$, producing alternating low- and high-ripple regions as overlap is avoided or occurs. When $n/k$ is an integer (e.g., $n=12$), the transition pattern repeats across segments and ripple depends strongly on $(\Delta d,k)$; when $n/k$ is not an integer (e.g., $n=5$), the transition pattern becomes asymmetric and thus the ripple becomes less sensitive to $k$.
}

    \label{fig:simulation2}
\end{figure}

\subsection{Non–Dimensional Analysis}
To generalize the model and facilitate comparison between different cam geometries and operating conditions, we conducted a non-dimensional analysis in which we express the system variables in non–dimensional form. The characteristic quantities used for normalization are the cam angular period $2\pi$, the maximum cam lift $H = \max(h)$, and the traction force limit $\mu N$. The non–dimensional parameters are therefore defined as follows:
\[
\tilde{\theta} = \frac{\theta}{2\pi}, 
\qquad
\tilde{h}(\tilde{\theta}) = \frac{h(\theta)}{H}, 
\qquad
\tilde{F}_{slider}(\tilde{\theta}) = \frac{F_{slider}(\theta)}{\mu N}.
\]

Substituting into the expression for the instantaneous slider force from Eq.~\eqref{eq:Fs_cases} and using the non–dimensional cam slope $\tfrac{d\tilde{h}}{d\tilde{\theta}}$, the normalized slider force becomes:
\begin{equation}\label{eq:nd_single_slider}
\tilde{F}_{slider}(\tilde{\theta}) = f(\lambda,\tilde{\theta})\,\mathrm{sign}\!\left[-\frac{d\tilde{h}}{d\tilde{\theta}}(\tilde{\theta})\right],
\end{equation}
where $f(\lambda,\tilde{\theta})$ is the previously introduced smoothing function that governs the transition between the forward and reverse friction limits. In the ideal case of saturating at maximum traction ($\mu N$) at all time, i.e. in case of sliders slipping at all time, the smoothing effect vanishes ($\lambda=0$), and $\tilde{F}_{slider}$ alternates between $\pm 1$. Increasing $\lambda$ produces smoother transitions corresponding to compliant or lubricated contact conditions.

The non–dimensional cycle–averaged thrust of a single slider can then be written as
\begin{equation}\label{eq:nd_{adv}vg_slider}
\bar{\tilde{F}}_{slider} 
= \frac{1}{2\pi}\int_{0}^{2\pi} \tilde{F}_{slider}(\tilde{\theta})\,d\tilde{\theta}
= (d_{ret} - d_{adv}) - \tilde{F}_{\mathrm{loss}},
\end{equation}
where $\tilde{F}_{\mathrm{loss}} = F_{\mathrm{loss}}/\mu N$ is the normalized internal loss. 

For a capsule employing $n$ uniformly phase–shifted sliders, the instantaneous non–dimensional thrust is given by
\begin{equation}\label{eq:nd_superposition}
\tilde{F}_{\mathrm{capsule}}(\tilde{\theta})
= \frac{1}{n}\sum_{i=1}^{n}\tilde{F}_{slider}\!\left(\tilde{\theta}-\frac{i-1}{n}\right),
\end{equation}
and the corresponding cycle–averaged non–dimensional thrust is:

\begin{equation}\label{eq:nd_{adv}vg_capsule}
\bar{\tilde{F}}_{\mathrm{capsule}} = (d_{ret} - d_{adv}) - \tilde{F}_{\mathrm{loss}}.
\end{equation}

This expression emphasizes that the non-dimensional average thrust depends only on the duty asymmetry $(d_{ret} - d_{adv})$ governed by the cam geometry and the loss term $\tilde{F}_{\mathrm{loss}}$ per slider.

\section{Simulation of OSCAR}\label{sec:simulation}
We used the analytical model in Section~\ref{sec:modeling} to simulate OSCAR’s kinematics and thrust under the ideal, friction-limited assumption that each slider’s traction saturates at $\mu N$. The simulation study aimed, first, to clarify how the three implemented cam profiles translate into slider motion and capsule thrust, and second, to examine how the cam duty fractions $(d_{\mathrm{ret}},d_{\mathrm{adv}})$, the jump count $k$, and the number of sliders $n$ collectively shape both the instantaneous thrust profile and its cycle-average.

\subsection{Kinematics and Thrust Force of Implemented Cam Designs}
We first simulated the single-slider kinematics for our three implemented cam designs (with $k=1,2,3$) by representing each cam through its lift function $h(\theta)$ and computing the slider displacement and velocity using Eqs.~(\ref{eq:camAngle})--(\ref{eq:Fs_cases}) (Fig.~\ref{fig:simulation1}(a)). Due to the kinematic constraint between the slider and the cam profile, the slider displacement directly follows $h(\theta)$. Accordingly, the normalized displacement $x_{slider}(\theta)/h_{max}$ (Fig.~\ref{fig:simulation1}(a)) exhibits one, two, and three rise--return motion reversals over one revolution for $k=1,2,3$, respectively. The corresponding normalized velocity profiles $\dot{x}_{s}(\theta)/(\omega h_{max})$ indicate how rapidly each stroke is executed and when the transitions between advance and retract occur (zero crossings) over the full $2\pi$ cycle. These transition angles define when a slider contributes forward versus backward traction (Fig.~\ref{fig:simulation1}(a)). In our three cam designs, we kept the advance-phase angle constant (red in Fig.~\ref{fig:simulation1}(a)), and therefore the peak advance (negative) velocity is identical across cams. Increasing $k$ thus compresses only the retract interval (green in Fig.~\ref{fig:simulation1}(a)), which increases the peak retract (positive) velocity.

Using these kinematics, we evaluated the normalized instantaneous traction of a single slider, $\tilde{F}_{\mathrm{slider}}(\theta)$, from Eq.~\eqref{eq:nd_single_slider} (Fig.~\ref{fig:simulation1}(a)). Under the friction-limited assumption (sliding always in slip), the traction saturates at $\pm 1$ (i.e., $\pm \mu N$ after normalization), with the sign set by the motion direction. We then obtained the normalized capsule thrust $\tilde{F}_{\mathrm{capsule}}(\theta)$ by superimposing the phase-shifted contributions of all sliders ($n=12$) using Eq.~\eqref{eq:nd_superposition}, and computed the cycle-average $\bar{\tilde F}_{\mathrm{capsule}}$ using Eq.~\ref{eq:nd_{adv}vg_capsule} (Fig.~\ref{fig:simulation1}(b)).

The computed instantaneous normalized capsule thrust $\tilde{F}_{\mathrm{capsule}}(\theta)$ shows two components (Fig.~\ref{fig:simulation1}(b)): a mean value and a small periodic ripple around that mean. The mean values for the three cams reflect the cycle-averaged thrust $\bar{\tilde{F}}_{\mathrm{capsule}}$, which is set by the duty asymmetry $\Delta d=d_\mathrm{ret}-d_\mathrm{adv}$ of each profile (see Fig.~\ref{fig:simulation2}(a)). The traces of the slider counts $n_{ret}$ and $n_{adv}$ over the cycle (Fig.~\ref{fig:simulation1}(b)) reveal the origin of the ripple. Specifically, the ripple arises from brief count changes during motion reversals when the slider spacing does not align exactly with the angular lengths of the advance and retract intervals. In that case, one slider switches phase while the preceding slider has not yet left it, producing a transient change in $n_{ret}$ and $n_{adv}$. This transient reduces temporarily the number of sliders contributing to forward thrust, resulting in the periodic ripple observed in $\tilde{F}_{\mathrm{capsule}}(\theta)$.

\subsection{Effect of Cam Duty Asymmetry, Jump Count, and Slider Count on Thrust}
To study the effects of the cam duty fractions $(d_{ret},d_{adv})$, the cam jump count $k$, and the number of sliders $n$, we simulated the capsule across a range of duty fractions, jump counts, and slider counts, and quantified their effects on both the cycle-averaged thrust and the periodic ripple in the instantaneous thrust profile.

As shown in Fig.~\ref{fig:simulation2}(a), the cycle-averaged normalized thrust $\bar{\tilde{F}}_{\mathrm{capsule}}$ is determined solely by the cam duty fractions $(d_{ret},d_{adv})$ and varies linearly with the duty asymmetry $\Delta d=d_{ret}-d_{adv}$ independent of the jump count $k$ and the number of sliders $n$. Because $\Delta d$ sets the net imbalance between the retract and advance intervals over a cycle, it directly controls both thrust direction and magnitude: a positive duty asymmetry $d_{ret}>d_{adv}$ ($\Delta d>0$) generates forward (positive) thrust, a negative duty asymmetry $d_{ret}<d_{adv}$ ($\Delta d<0$) generates backward (negative) thrust, and a null duty asymmetry yields zero mean thrust (Fig.~\ref{fig:simulation2}(a)). This highlights duty asymmetry as a potential handle for bi-directional propulsion.

To quantify the periodic ripple in the normalized instantaneous thrust $\tilde{F}_{\mathrm{capsule}}(\theta)$, we define the ripple as the coefficient of variation over one cycle, i.e., the standard deviation normalized by the cycle-average, $\sigma_{\tilde{F}_{\mathrm{capsule}}}/\bar{\tilde{F}}_{\mathrm{capsule}}$. As explained earlier, this ripple arises from overlapping slider phase switches, which temporarily change the number of sliders in each phase (as in Fig.~\ref{fig:simulation1}(b)). The duty asymmetry $\Delta d$, jump count $k$, and slider count $n$ therefore set when and how often these count changes occur.

Specifically, $\Delta d$ sets the angular lengths of the retracting and advancing intervals within each of the $k$ segments of the cycle, and therefore sets the phase boundaries (advance vs.\ retract) in each segment. As $\Delta d$ is varied for a fixed $k$, these boundaries slide relative to the slider spacing $2\pi/n$, changing whether sliders switch phases sequentially (one-by-one) or with temporal overlap. As a result, the ripple (Fig.~\ref{fig:simulation2}(b)) is governed by the interplay between the geometric slider spacing ($2\pi/n$) and the cam timing $(k,\Delta d)$, producing alternating low- and high-ripple regions as $\Delta d$ is swept for each $k$.

The role of $k$ is to set how many rise--return sequences occur per revolution, and thus how many phase switches occur per cycle. When the $n$ sliders are evenly distributed across the $k$ segments (i.e., $n/k$ is an integer), the same transition pattern repeats from segment to segment, so overlap events recur at the same relative times and generate ripple that depends on the combination of $\Delta d$ and $k$; as reflected in the $n=12$ case across $k=1,2,3$ (bottom-right plot and heat maps in Fig.~\ref{fig:simulation2}(b)). In contrast, when $n/k$ is not an integer, sliders are not equally distributed across segments, so overlaps occur at different times in different segments and do not repeat consistently; consequently, the ripple becomes less sensitive to $k$, as illustrated by the identical curves for $n=5$ (bottom-left plot in Fig.~\ref{fig:simulation2}(b)).

\subsection{Design Implications from Simulation}
The simulations indicate that mean thrust and thrust smoothness are tuned by different parameters. The cycle-averaged thrust is set solely by the duty asymmetry $\Delta d=d_{ret}-d_{adv}$: it scales linearly with $\Delta d$ and reverses sign with it, enabling bi-directional control through cam timing. In contrast, the jump count $k$ and slider count $n$ mainly affect thrust smoothness (ripple) by shaping (together with the duty asymmetry $\Delta d$) whether slider phase switches overlap during reversals.

\section{Experimental Validation}\label{sec:experiment}
\subsection{Force Characterization Experiment}
To characterize the thrust force generated by OSCAR, we conducted two complementary sets of experiments using \textit{ex-vivo} porcine colon tissue. The first quantified the force cycle of a single slider, allowing us to analyze the contact mechanics at the single slider level. The second measured the net thrust force of the complete capsule operating inside an intact \textit{ex-vivo} colon segment, to connect the slider-level mechanics to the whole-system performance, and validate our theoretical model. The full-robot tests also enabled exploration of how actuation speed and cam design influence the generated thrust.

\begin{figure}
    \centering
    \includegraphics{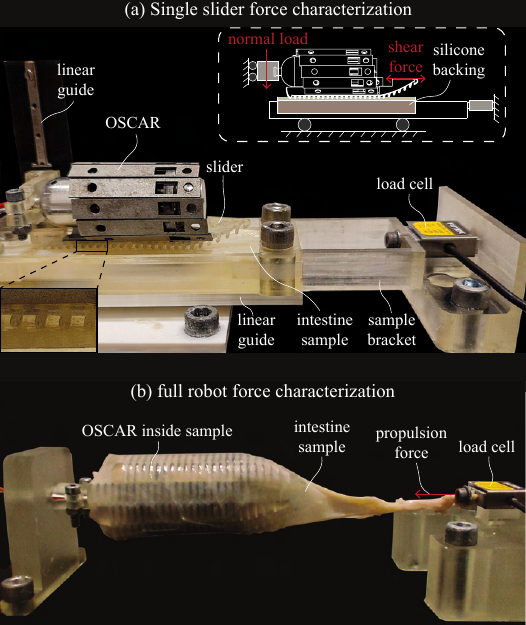}
    \caption{\textbf{Experimental setups for characterizing the thrust force generated by OSCAR.} (a) Single-slider force characterization. An \textit{ex-vivo} porcine colon sample is opened and mounted flat on a silicone-backed bracket, which is guided horizontally and instrumented with a single-axis load cell to measure the traction force from the tissue side. The robot is mounted on a vertical guide so that a single slider presses against the tissue under controlled normal loading. This configuration enables direct observation of the stick--slip traction cycle and captures the viscoelastic response of the wall, which is essential for understanding the slider--tissue contact mechanics. (b) Full-robot thrust characterization. OSCAR is inserted coaxially into an intact porcine colon segment attached to a single-axis load cell. Measuring traction from the tissue side again allows tacking of the traction force on the wall which includes its viscoelastic response, linking individual-slider mechanics to whole-robot propulsion. This setup is further used to evaluate how actuation speed, cam duty cycle, and the number of motion reversals per cycle (cam jumps) influence the net thrust generated by the capsule.}
    \label{fig:setup}
\end{figure}

For the single-slider experiment (Fig.~\ref{fig:setup}a), we cut open an \textit{ex-vivo} porcine colon sample and mounted it flat on a bracket with a silicone-backing layer underneath. We positioned the bracket horizontally on a linear guide and attached it to a co-axial single-axis load cell to measure the tangential forces acting on the tissue. OSCAR was mounted on a vertical guide so that only a single slider was pressed downward onto the tissue under a controlled normal load. Measuring the traction from the tissue side was essential to capture the viscoelastic response of the tissue, which is critical to understand the physics of the slider-tissue interaction. Before each trial, we hydrated the tissue surface with saline and reset the baseline force by lifting and reloading the robot. We repeated each condition six times.

For the full-capsule experiment (Fig.~\ref{fig:setup}b), we attached an intact porcine colon segment to a single-axis load cell at one end and inserted the robot coaxially from the other. We fixed the capsule from its base into a mounting bracket to maintain alignment with the load-cell axis. As in the single-slider setup, measuring traction from the tissue side was essential for capturing the wall’s viscoelastic response and linking the slider-level mechanics to whole-system behavior. Using this configuration, we evaluated how actuation speed, cam duty cycle, and the number of reversals (i.e., cam jumps) influence the generated thrust force. Before each run, we hydrated the colon specimen externally with saline, and after each trial we relaxed the tissue to release any tension accumulated during the trial. Each experimental condition was performed in six repeated trials.

\subsection*{Proof-of-Concept Locomotion Experiment}
In this proof-of-concept experiment, we evaluated the capsule’s ability to traverse colonic tissue using \textit{ex-vivo} porcine colon samples. Testing in real tissue is essential for validating friction-based locomotion systems. Unlike silicone phantoms, which typically exhibit elevated and highly uniform friction, \textit{ex-vivo} colon provides a more physiologically representative mechanical properties including viscoelastic and frictional properties which are of great influence on the performance of the capsule as indicated by the model. 

We prepared six \textit{ex-vivo} intestinal tissue samples, each approximately $1~\mathrm{m}$ in length, and immersed them in saline prior to each trial to ensure surface lubricity. The capsule was then introduced at the proximal end of the segment to ensure full wall contact. To impose a more stringent locomotion challenge, the intestinal segments were left in their natural, deflated configuration—without insufflation or suspension—requiring the capsule not only to generate forward thrust but also to pull itself through self-contacted, folded tissue. After activating the capsule, we recorded its progression along the sample until the end of each run.

At the end of every trial, the inner wall of the intestine samples was inspected under a microscope to observe any potential visual tissue damage associated with capsule–wall interaction.

\section{RESULTS AND DISCUSSION}\label{sec:discuss}
\subsection{Single Slider Stick-Slip Contact Mechanics}
\begin{figure}[t!]
    \centering
    \includegraphics[width=\linewidth]{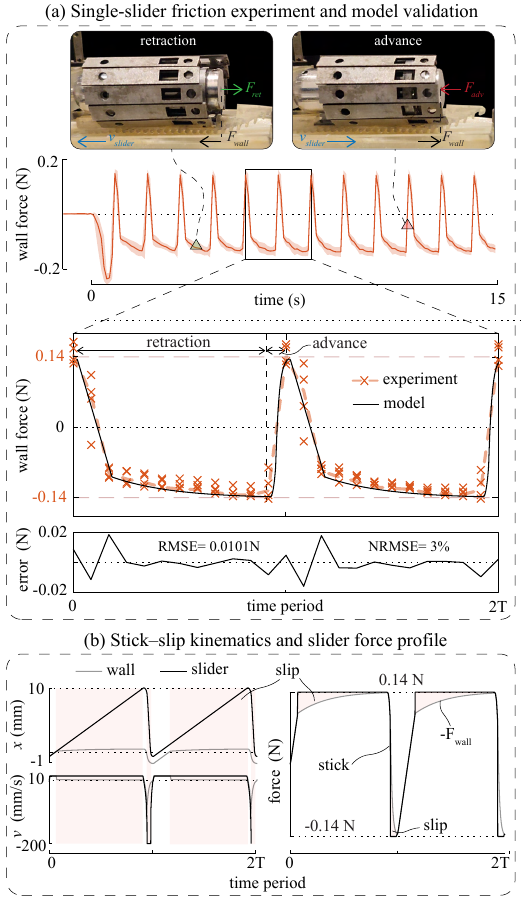}
    \caption{\textbf{Single–slider friction experiment and validation of the viscoelastic stick–slip model.} (a) Experimental setup and measured wall–force response during alternating retraction and advance of a single slider. The recorded traction force exhibits symmetric peak traction ($\pm 0.14$~N) and clear stick–slip transitions shaped by the wall’s viscoelastic response. The zoomed-in view of the traction force profiles show the linear build-up during stick and the exponential convergence toward the friction limit during slip, together with the model prediction obtained using the experimentally-identified elongation stiffness $k_{tissue} \approx 130 \mathrm{N.m^{-1}}$ and viscous damping $c_{tissue} \approx 6 \ \mathrm{N.s.m^{-1}}$ parameters. The bottom panel shows error profile over two periods, with the model achieving $\mathrm{RMSE}=0.0101$~N and $\mathrm{NRMSE}\approx 3.6\%$. (b) Estimated wall and slider kinematics and the corresponding stick–slip force evolution. The model reconstruction shows how slider and wall motion coincide during stick, producing linear traction growth, and how the slider force saturates during slip while the wall relaxes according to the tissue’s viscoelastic dynamics. These results demonstrate that the stick-slip formulation accurately reproduce the recorded interaction and reveal the underlying contact mechanics. Refer to Supplementary video.2 for a video demonstration of the single slider force characterization experiment.}
    \label{fig:single_slider}
\end{figure}

The recorded tangential force of a single slider against the colonic tissue (Fig.~\ref{fig:single_slider}(a)) revealed their contact mechanics. As the cam drives the slider through the alternating phases of retraction and advance, the measured force switches between positive and negative traction with nearly symmetric peak magnitudes ($\pm 0.14~\mathrm{N}$), which is expected thanks to the symmetric surface design of the slider. The durations of the retraction and advance phases constitute $\approx 91.1\%$ and $\approx 8.5\%$ of the time period, respectively. These duration percentages yield a duty asymmetry of $\Delta d \approx 0.82$, perfectly matching the duty-cycle asymmetry of the single-jump cam used in this experiment, confirming that the slider is indeed kinematically constrained to the cam profile.

Examining the measured force profile in Fig.~\ref{fig:single_slider}(a) reveals distinct behaviors in the two motion phases. During retraction, the traction on the wall decreases (increases from the capsule's perspective) linearly up to a threshold of $\approx -0.08~\mathrm{N}$, beyond which the force slowly converges to the retraction friction limit $-0.14~\mathrm{N}$ following an exponential decay. In contrast, the advance phase exhibits an approximately linear increase (decrease from the capsule's perspective) in traction until the advance friction limit $0.14~\mathrm{N}$ (negative thrust from the capsule's perspective) is reached. These smooth reversals between peak positive and negative traction reflect the viscoelastic response of the colonic wall, whose deformation and time-dependent relaxation shape the contact mechanics. Since the force is measured from the tissue wall side (Fig.\ref{fig:setup}(a)), this viscoelastic behavior is fully captured in the experimental data.

The observed contact behavior (linear followed by exponential convergence in retraction and linear in advance) can be interpreted through the mechanics of stick and slip. During stick, the slider and the colonic wall move together, with no relative velocity ($v_{\text{rel}} = 0$), and therefore the tissue is kinematically constrained to follow the slider displacement. As the slider moves, the wall stretches tangentially, building up an internal viscoelastic force governed by the elastic and viscous properties of the tissue. In this displacement-driven regime, the tangential traction experienced by the slider is equal and opposite to the elastic restoring force of the tissue, producing an approximately linear increase or decrease in traction depending on the slider motion. Stick persists until this internal tissue force reaches the static friction limit ($F_{tissue}>=\mu_{s}N$).

Once the accumulated internal force exceeds the static friction threshold, the slider force can no longer stretch the tissue wall further, and thus, the slider starts to slip relative to the wall ($|v_{\text{rel}}| > 0$). In this regime, the slider thrust force saturates at the kinetic friction limit $\mu_{k}N$, while the tissue is no longer kinematically constrained to the slider. Instead, its internal viscoelastic state evolves under a force-driven condition. The tissue relaxes from its previously stored deformation toward the equilibrium set by the driving kinetic friction force. This relaxation follows an exponential decay with a characteristic time constant governed by the viscoelastic properties of the tissue, producing the gradual convergence toward the friction limit observed in the measurements. As a result, the wall displacement and internal force evolve continuously during slip, while the slider thrust force remains fixed at the kinetic friction limit. Slip continues until the internal tissue force becomes smaller than the kinetic friction limit ($F_{tissue}<\mu_{k}N$), after which contact returns back to the stick regime. This stick--slip dynamics can be mathematically described as follows:
\begin{equation}
F_{\text{slider}} =
\begin{cases}
-\,F_{\text{tissue}}, 
&|F_{\text{tissue}}| < \mu_s N, \text{(stick)} \\[4pt]
\mu_k N\,\mathrm{sgn}\!\big(v_{\mathrm{rel}}\big), 
&|F_{\text{tissue}}| \ge \mu_k N, \text{(slip)}
\end{cases}
\label{eq:stick_slip_law}
\end{equation}
where $F_{\text{tissue}}$ is the internal force of the tissue, $N$ is the normal load, $\mu_s$ and $\mu_k$ are the static and kinetic coefficients of friction, respectively, and $\mathrm{sgn}(v_{\mathrm{slider}})$ is the sign of the slider velocity.

To capture the viscoelastic response of the colonic wall during both stick and slip, we model the tangential tissue mechanics as a single in-parallel spring--dashpot element, also known as Kelvin--Voigt model \cite{Gregersen2003, Siri2020Macro, Kim2006TL}, characterized by an effective elongation stiffness $k_{tissue}$ and damping coefficient $c_{tissue}$. During stick, the wall is kinematically constrained to follow the slider, and thus the internal force of the tissue can be obtained directly from the viscoelastic resistance to motion, expressed as follows:
\begin{equation}\label{eq:stick}
F_{\text{tissue}}(t) = k_{tissue}\,x_{wall}(t) + c_{tissue}\,\dot{x}_w(t), \ \text{(stick)}
\end{equation}

where $x_{wall}(t)$ is the displacement of the wall. Once slip initiates and the interface becomes force-driven, the tissue is no longer constrained by the slider kinematics. Therefore, its internal force, and consequently its deformation, relaxes toward the friction-limited driving force $\mu_k N\,\mathrm{sgn}(v_{\mathrm{slider}})$. This relaxation in force and displacement can be captured with the Kelvin--Voigt first-order relaxation law as follows:
\begin{equation}
\begin{aligned}
\dot{F}_{\text{tissue}}(t) &= -\frac{1}{\tau(v)}\,\big(F_{\text{tissue}}(t) - F_{\text{slider}}(t)\big),\\[6pt]
\dot{x}_{w}(t) &= -\frac{1}{\tau(v)}\!\left(x_{w}(t) - \frac{F_{\text{tissue}}(t)}{k_{tissue}}\right), \ \text{(slip)}
\end{aligned}\label{eq:slip}
\end{equation}
with $\tau$ representing the effective relaxation time constant of the tissue. 
The elongation stiffness $k_{tissue}$ and viscous coefficient $c_{tissue}$ of the tissue wall were identified directly from the single--slider force recordings (Fig.\ref{fig:single_slider}(a)) by fitting the stick portions of the cycle. In this displacement-driven regime, the measured traction follows the Kelvin--Voigt resistance to the imposed wall motion, and thus the linear traction build-up during stick provided $k_{tissue}$ and $c_{tissue}$ in Eq.~\eqref{eq:stick}, which we identified to be $\approx 130~\mathrm{N\,m^{-1}}$ and $\approx 6~\mathrm{N\,s\,m^{-1}}$, respectively, in both phases of advance and retraction.

The exponential convergence toward the kinetic-friction limit during slip then provided the effective relaxation time constant $\tau(v)$ in Eq.~\eqref{eq:slip}. The convergence showed a rate-dependent relaxation time constant $\tau(v)$, yielding $\tau \approx 0.02~\mathrm{s}$ during advance and $\tau \approx 0.45~\mathrm{s}$ during retraction, consistent with previously reported rapid relaxation rates of intenstinal tissue at higher strains~\cite{Kim2006TL,Tan2011}. Additionally, the faster relaxation time $\tau$ at higher actuation speed (advance) is consistent with previously reported rate-dependent viscoelastic relaxation in intestinal tissue~\cite{Zhang2012,Ehteshami2015,Barducci2020}. 

Using the experimentally identified viscoelastic parameters together with the established stick--slip formulation detailed in Algorithm~\ref{alg:stickslip}, we were able to reproduce, with high fidelity, the measured stick-slip behaviour of the single-slider interaction with the tissue wall. The model captures the key qualitative features of the force response, including the linear force build-up during stick, the smooth transitions into slip, and the characteristic viscoelastic relaxation observed experimentally. Quantitatively, the model achieved a root-mean-square error of $\mathrm{RMSE}=0.0101~\mathrm{N}$ across cycles, corresponding to a normalized error of $\mathrm{NRMSE}\approx 3.62\%$ of the full force range, as shown in Fig.~\ref{fig:single_slider}(a), which is below the typical range reported for viscoelastic soft-tissue models, where residuals of $5-15\%$ are common due to inherent biological variability \cite{Gregersen2003}. This close agreement confirms that the dominant physics of the slider--tissue interaction are well explained by our viscoelastic stick-slip formulation, validating its utility for the multi-slider thrust model.

With the stick–slip formulation and viscoelastic parameters validated against the measured force profile, we can now reconstruct the full kinematics of the wall and the corresponding forces acting on both the slider and the tissue. As shown in Fig.~\ref{fig:single_slider}(b), the model shows how the wall displacement follows the slider displacement during stick, producing the linear traction build-up, and how it subsequently relaxes during slip as the slider force saturates at the friction limit. During stick, the slider and wall forces follow one another exactly, reflecting their kinematic constraint, whereas during slip the slider force remains fixed at the kinetic friction limit while the tissue wall force relaxes according to the tissue’s viscoelastic relaxation. This mechanics reconstruction provides a clear picture of the local stick-slip interaction at the level of a single slider, providing the basis to understand the collective behavior of the capsule.

\subsection{Capsule Thrust Mechanics}
\begin{figure*}
    \centering
    \includegraphics[width=\linewidth]{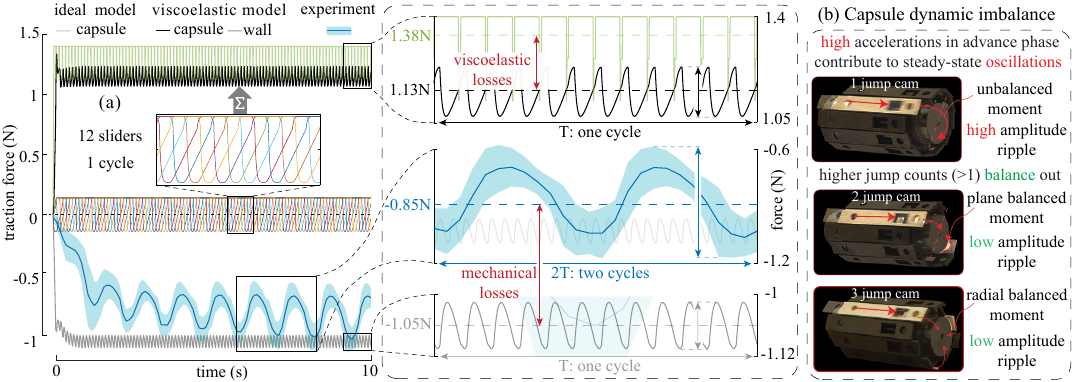}
    \caption{\textbf{Capsule thrust force and dynamic imbalance.}
(a) Comparison between the measured traction force generated by the capsule on the \textit{ex-vivo} colon wall (blue), the wall and capsule thrust force estimated by the viscoelastic model (grey and black), and the capsule thrust force estimated by the ideal model (the friction-limited case) (light green). The ideal model predicted cycle-averaged capsule thrust force ($1.38~\mathrm{N}$) compared to ($1.13~\mathrm{N}$) predicted by the viscoelastic model, which reflects the losses due to wall viscoelasticity. The absolute value of the experimentally measured steady-state average traction ($\approx 0.85~\mathrm{N}$) is less than  the absolute value of the wall force predicted by the viscoelastic model ($1.05~\mathrm{N}$), attributed to the mechanical losses in the system. Insets show model–experiment comparison over one and two cam cycles, highlighting viscoelastic losses (top) and system-level mechanical losses (bottom).  
(b) Illustration of the inertial moment imbalance generated by sliders during the rapid advance phase. In the single-jump cam, only one slider undergoes the high acceleration motion reversal when switching to advance phase without any balancing counter force, producing a net rocking moment on the capsule and a large-amplitude ripple in the measured thrust. For higher jump counts, sliders advance in symmetric groups (two opposite sliders in the two-jump cam, three evenly spaced sliders in the three-jump cam), balancing the inertial moments and substantially reducing the thrust ripple. Refer to supplementary video.3 for a video demonstration of the full capsule force characterization experiment.}
    \label{fig:CapsuleThrustDynamics and Effects}
\end{figure*}

Building on the validation of the single-slider contact mechanics, we evaluated the traction of the full capsule using the same (single jump) cam profile, allowing for a direct comparison between the measured thrust and the predictions of the validated stick-slip model. The full capsule force experiment showed a mean steady-state traction force on the tissue wall of $\approx -0.85~\mathrm{N}$ (see Fig.~\ref{fig:CapsuleThrustDynamics and Effects}(a)), confirming the capsule's ability to generate steady thrust. The measured steady-state wall traction deviates from the $-1.05~\mathrm{N}$ predicted by the viscoelastic model, attributed to the mechanical losses associated with the full system assembly, which are not accounted for in the model. Additionally, the corresponding capsule thrust force computed using our viscoelastic model was $1.13~\mathrm{N}$, compared to $1.38~\mathrm{N}$ predicted by the model in the ideal case of the contact force always saturating at the kinetic friction limit. The deviation in this case arises from the viscoelastic properties which alter the force profile waveform, yielding a different net cycle-averaged thrust.

The recorded traction force on the wall also showed steady-state oscillations (ripple) with an amplitude of $\approx 0.3~\mathrm{N}$, as shown in Fig.~\ref{fig:CapsuleThrustDynamics and Effects}(a). While our model predicts that the capsule traction force, and consequently the force on the tissue wall, will exhibit oscillations stemming from the instantaneous overlapping of sliders switching from retract to advance motion, the experimental data diverge from these predictions. The observed amplitude was six times greater than the ripple amplitude predicted by the viscoelastic model. Furthermore, the experimental frequency matched the actuation frequency (one peak per cycle), whereas the model predicted a frequency of $12$ peaks per cycle, corresponding to one motion reversal of each of the twelve sliders per one cycle. This mismatch in both amplitude and frequency suggests that the source of the observed ripple in the experiment differs from the mechanism assumed in the model.

The unexpectedly large ripple in the full-capsule force can be explained by a small rocking moment during the steep advance phase. Although the slider is kinematically constrained to the cam, its rapid acceleration produces an inertial force with an axial component, which generates a tilting moment on the capsule (Fig.~\ref{fig:CapsuleThrustDynamics and Effects}(b)) and induces a small rotation $I\ddot{\phi}=M$. Even sub-degree tilts can create axial displacements that, with the measured wall stiffness $k_{\mathrm{tissue}}$, lead to force oscillations consistent with the $\sim0.6$~N peak-to-peak ripple. The effect is strongest for the single-jump cam due to the unbalanced rocking moment, and is reduced for higher jump-count cams where multiple sliders simultaneously advance, which balance the capsule and partially cancel out the rocking moment moment (Fig.~\ref{fig:CapsuleThrustDynamics and Effects}(b)).

\begin{algorithm}[t]
\caption{Stick--slip capsule-wall interaction model}
\label{alg:stickslip}
\begin{algorithmic}[1]

\Require sampled time series $t(n)$, $\Delta t$, $x_{slider}$, $k_{tissue}, c_{tissue}, \mu_s,\mu_k, N, \tau_{ret}, \tau_{adv}$

\State $mode \gets \textsc{Stick}$

\For{$i = 2$ to $n$}
    \State $\Delta x_{slider} \gets x_{slider}(i) - x_{slider}(i-1)$;\quad
           $v_s \gets \Delta x_{slider} / \Delta t$
    \State $\tau \gets \tau_{ret}$ \textbf{if} $v_s \ge 0$ \textbf{else} $\tau_{adv}$;\quad
           $\alpha \gets 1 - e^{-\Delta t/\tau}$

    \If{$mode = \textsc{Stick}$}
        \State $x_{wall}(i) \gets x_{wall}(i-1) + \Delta x_{slider}$
        \State $v_{wall}(i) \gets \big(x_{wall}(i) - x_{wall}(i-1)\big)/\Delta t$
        \State $F_{wall}(i) \gets k_{tissue} x_{wall}(i) + c_{tissue} v_{wall}(i)$
        \If{$|F_{wall}(i)| \ge \mu_s N$};\ $mode \gets \textsc{Slip}$\ \EndIf

    \Else{\ $\textbf{if} \ mode = \textsc{Slip}$}
        \State $F_{\mathrm{target}}=\mu_k N\ \mathrm{sign}(v_{wall}(i))$ 
        \State $F_{wall}(i) \gets F_{wall}(i-1) + \alpha\big(F_{\mathrm{target}} - F_{wall}(i-1)\big)$
        \State $x_{\mathrm{target}}=F_{wall}(i)/k_{tissue}$
        \State $x_{wall}(i) \gets x_{wall}(i-1) + \alpha\big( x_{\mathrm{target}}- x_{wall}(i-1)\big)$
        \State $v_{wall}(i) \gets \big(x_{wall}(i) - x_{wall}(i-1)\big)/\Delta t$
        \If{$\big|F_{wall}(i)\big| < \big|k_{tissue} x_{wall}(i)\big| $};\ $mode\gets\textsc{Stick}$ \ \EndIf 
    \EndIf; \quad $F_{elastic}(i) \gets x_{wall}(i)k_{tissue}$
\EndFor;\quad \Return $x_{wall}, v_{wall}, F_{wall},F_{elastic}$

\end{algorithmic}
\end{algorithm}

\subsection{Influence of Actuation Speed and Duty Asymmetry}

\begin{figure}
    \centering
    \includegraphics[width=\linewidth]{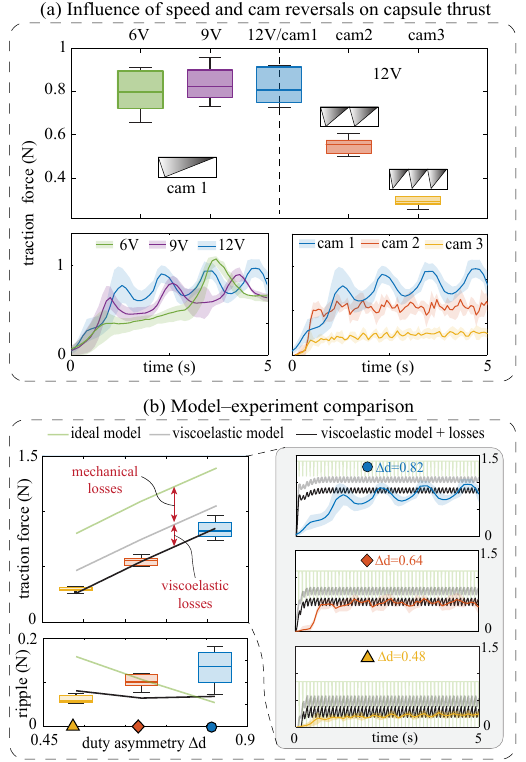}
    \caption{\textbf{Effect of speed and cam jumps on capsule thrust, and comparison between models and experiments.}
(a) Cycle-averaged traction forces generated by OSCAR under different voltages ($6\mathrm{V}, 9\mathrm{V}, 12\mathrm{V}$) corresponding to different motor speeds using the single-jump cam, and under the three different cam geometries with ($1,2,3$) cam jumps at $12\mathrm{V}$ driving voltage. Box plots and the corresponding time-evolution traces show the mean thrust remained approximately constant across the different speed, whereas increasing the number of cam jumps reduces the net traction and produces progressively smoother force profiles due to improved moment balancing across sliders. (b) Comparison between measured traction, the ideal friction model, the viscoelastic model, and the viscoelastic model with subtracted mechanical losses, plotted against duty-cycle asymmetry $\Delta d$. 
Viscoelasticity lowers the ideal thrust, and additional mechanical losses observed in the full system shifts the prediction further downward, closely matching the experimental data. 
The lower-left panel shows corresponding ripple amplitudes. 
Right: the corresponding time-evolution traces for each cam design demonstrate how model and experiment agree in mean thrust while differing in ripple characteristics due to system-level inertial effects.}
    \label{fig:speedAndCam}
\end{figure}

Examining the role of the actuation speed on thrust revealed that the generated thrust remained effectively unchanged across the input voltages tested. As shown in Fig.~\ref{fig:speedAndCam}(a), operating the motor at 6~V, 9~V, or 12~V, corresponding to three different actuation speeds, produced nearly identical steady–state averaged traction forces of $\approx 0.85$~N, with the time–evolution profiles closely overlapping, indicating that the capsules' thrust is speed-independent. This speed independence confirms the validity of the Coulomb friction law, used in the theoretical model, within the operating speeds tested. Furthermore, it shows that the generated thrust force is robust against speed variations that can result from varying normal loading.  

Modifying the cam geometry produced systematic and predictable changes in the mean thrust. Increasing the number of motion reversals (jumps) from one to two and three jumps decreased the cycle-averaged traction force, as shown in Fig.~\ref{fig:speedAndCam}(a). This decrease follows directly from the reduced duty asymmetry $\Delta d$ from $0.82$ to $0.64$ and $0.48$ of the single, double, and triple jump(s) cams, respectively. As the duty asymmetry decreases, a larger fraction of the sliders spend more time in the advance phase than in retraction, thereby reducing the generated net thrust. This trend is consistent with the analytical predictions of Eq.~\ref{eq:averageThrust} as depicted in Fig.~\ref{fig:simulation2}(a), in which the model anticipates a monotonic dependence of the averaged thrust force on the duty-cycle asymmetry. As shown in Fig.~\ref{fig:speedAndCam}(b), the model closely matches the recorded traction force once both viscoelastic effects and system-level mechanical losses are incorporated.

However, while the model matches the trend of the recorded average traction force, the observed ripple in the recorded traction waveform does not follow the model’s prediction. The model predicts the largest ripple amplitude for the three-jump cam and the smallest for the single-jump cam, due to the increased number of slider phase switching as the jump count increases. Experimentally, the opposite occurred: as shown in Fig.~\ref{fig:speedAndCam}(b), the single-jump cam exhibited the highest ripple amplitude, whereas the two- and three-jump cams showed progressively less ripple amplitude in their recorded traction force profiles. This discrepancy can be explained by the dynamic imbalance of the full capsule, depicted in Fig.~\ref{fig:CapsuleThrustDynamics and Effects}(b). The effect is expected to be most pronounced for the single-jump cam, and reduced for the two- and three-jump cams as the advancing-slider motion becomes more evenly distributed and balanced around the circumference of the capsule.  

The close qualitative and quantitative agreement between the model and the experiments in traction-force behavior across both speed variations and cam geometries supports the theoretical conclusion that OSCAR’s self‑propulsion is governed primarily by duty‑cycle asymmetry and the viscoelastic stick–slip behavior at each slider–wall interface, independent of speed. Additionally, the ripple observed in the experiments underscores the importance of phase distribution and synchronization of sliders, and the dynamic balance of the capsule's motion cycle in achieving smooth traction force.

\begin{figure*}
    \centering
    \includegraphics[width=\linewidth]{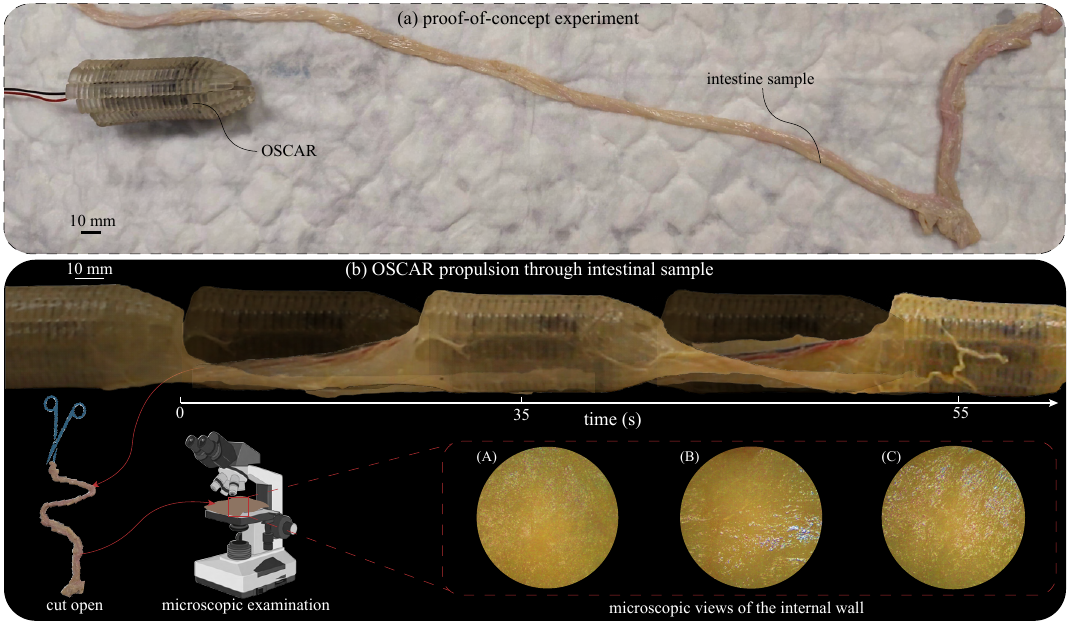}
    \caption{\textbf{\textit{ex-vivo} validation of locomotion and tissue safety in porcine intestine.}
(a) Proof-of-concept experimental setup showing OSCAR alongside a long, excised porcine colon sample in its natural, collapsed configuration (without insufflation).
(b) Time-lapse visualizing the progression of the capsule through the colon sample over a 55-second interval. The robot successfully propelled forward through the colon samples, achieving an average speed of $3.08~\mathrm{mm/s}$. This speed corresponds to a theoretical full-length colon traversal time of $\approx 8$~minutes, comparable to the standard for conventional colonoscopy.
Following locomotion, the intestinal tract was dissected (schematic inset) to assess the integrity of the tissue wall.
(A--C) Representative microscopic views of the internal mucosal wall at distinct locations along the traversal path. The images reveal an intact epithelial surface with no visible abrasions, hematomas, or disruption, confirming the atraumatic nature of the ovipositor-inspired self-propelling mechanism. Refer to supplementary video.4 for video demonstration of the locomotion validation experiment.}
    \label{fig:locomotion}
\end{figure*}

\subsection{Proof-of-Concept Locomotion Demonstration}
To demonstrate the self-propelling capability of OSCAR, we conducted locomotion validation experiments using \textit{ex-vivo} porcine colon samples, in which we inserted the capsule into intact samples of the \textit{ex-vivo} porcine colon and activated the capsule, as shown in Fig.~\ref{fig:locomotion}(a). Upon activation, the capsule successfully generated sufficient thrust to advance forward, effectively reopening the naturally collapsed lumen as it progressed (Fig.~\ref{fig:locomotion}(b)). Across six samples, OSCAR successfully traversed the full $100~\mathrm{cm}$ length at an average speed of $3.08 \pm 0.08~\mathrm{mm/s}$. Extrapolating this velocity to the average human colon length of $1.6~\mathrm{m}$ suggests a potential total intubation time of approximately eight minutes, a duration comparable to that of conventional colonoscopy~\cite{kim2021cecal}. Furthermore, post-trial microscopic inspection of the mucosal surface showed no visible damage to the tissue wall (Fig.~\ref{fig:locomotion}(b)). This demonstration highlights the potential of OSCAR to achieve clinically relevant locomotion speeds while maintaining safe interaction with the colon wall.

\subsection{Implications on Future Designs and Generality of Results}
While this study focused on introducing OSCAR and validating its working principle, the theoretical insights gained here provide general principles that go beyond the specific design of OSCAR.

On a system level, the duty-cycle framework used in this work to describe the retraction and advance phases of the cam profile is a robust, device-agnostic tool. It shows that locomotion fundamentally depends on the asymmetry between the ``grip'' and ``release'' phases, regardless of the actuation source. This abstraction therefore provides a transferable metric for future designs, whether driven by cams, linear actuators, or soft robotic transmissions.

On an interface level, the stick--slip model presented in Algorithm~\ref{alg:stickslip} sheds light on what constitutes an ideal capsule design for systems that rely on sliding friction to generate locomotion. In this context, the ideal sliding-friction capsule is one in which sliders slip for most of the cycle, making thrust less sensitive to wall viscoelasticity. According to the stick--slip model, this can be achieved by increasing the stroke height of the slider (resulting in higher elastic resistance $k.x_{slider}$ of the wall) and/or by increasing slider velocity $v_{\mathrm{slider}}$ through steeper motion profiles (resulting in higher viscous resistance  $c.v_{slider}$ of the wall), both of which help the sliders break the stick phase and overcome the viscoelastic foce of the wall, enabling quick transition to the slip mode. For a sliding-friction capsule to self-propel, it must be able to transition from stick to slip during its motion cycle. If the capsule remains in stick mode, the colon wall will stretch back and forth with the sliders, trapping the capsule and thus preventing it from propelling forward.

Finally, the stick--slip model presented in this study provides a general tool to simulate capsule designs that rely on sliding friction to generate thrust. The model takes the slider system kinematics and tissue viscoelastic properties as inputs, and predicts the capsule--wall interaction, the resulting capsule thrust, and the associated wall deformation and internal forces. Since we validated the model experimentally, it can be used to guide future iterations of OSCAR and other sliding-friction capsule designs by enabling simulation-based design optimization and performance prediction prior to implementation.
\section{Limitations and Future Directions}
While the presented results demonstrate OSCAR’s ability to self-propel and its underlying working principle, several limitations and challenges remain to be addressed to optimize the design and performance of OSCAR, and translate it into a fully operational clinical platform. 

The current capsule prototype was designed to validate the locomotion working principle and study the factors that affect its performance, resulting in a self-contained architecture with limited available internal volume, which limited the ability to integrate essential clinical payloads such as a camera and biopsy tools. Future iterations will address this constraint by potentially relocating the motor outside the capsule and transmitting motion through a flexible drive shaft, thus freeing internal volume for tool integration. Additional space could also be gained by reducing the number of sliders. Although this would reduce the thrust capacity of the capsule, the loss could be compensated for by integrating active friction-control methods, such as ultrasonic friction modulation~\cite{atalla2024UFM} or electroadhesion \cite{Guo2020}, to control the friction of individual sliders with the wall, which could potentially increase the effective friction anisotropy, and thus increase the average thrust.

The current cam mechanism generates unidirectional forward propulsion. While sufficient for cecal intubation, clinical workflows require bidirectional propulstion for better controllability and fine inspection during withdrawal. Future designs will incorporate a switchable or dual-profile cam system to enable bi-directional locomotion. Additionally, geometric adaptability is required to handle the high variability of diameters and shapes along the different segments of the colon. The integration of radially compliant or expandable sliders, similar to those proposed in~\cite{Consumi2023,Atalla2024MechanicallyInflatable}, would allow the capsule to maintain optimal wall contact pressure across varying lumen sizes and shapes.


Another limitation of the current study is that all experiments were performed under ambient conditions in straight \textit{ex-vivo} colon segments. \textit{In-vivo}, the colon is highly tortuous, variable in size and shape, with varying boundary conditions along its length and subject to intra-abdominal pressure, all of which can influence the capsule’s thrust capacity. For example, the intra-abdominal pressure acting on the colon wall can potentially increase the normal contact force between the capsule and the wall, which can both increase the frictional grip and the viscoelastic damping of the wall. Additionally, curved segments can introduce additional drag, particularly through the tether of the capsule which becomes increasingly significant as the capsule goes deeper into the colon. Future work will therefore extend both the experiments and the analytical model to evaluate OSCAR's thrust capacity and drag in curved, pressurized, and geometrically-constrained conditions.

Finally, from a sterility perspective, the current slider system includes small gaps between neighboring sliders, which may trap biological residue, posing clinical risks and sterilization challenges. To satisfy clinical sterilization standards, future designs will explore encapsulating the mechanism within a continuous, deformable medical-grade skin, eliminating the existing gaps between sliders and potentially improving sterility. Such a continuous skin could potentially also enhance the self-propelling performance of the capsule by increasing the effective movable contact area with the mucosa. 

\section{Conclusion}\label{sec:conclusion}
We introduced OSCAR, an ovipositor-inspired self-propelling capsule robot designed to overcome the fundamental traction limitations of soft-tissue locomotion and reduce the discomfort of conventional colonoscopy by mimicking how parasitic wasps coordinate their ovipositor valves to transport eggs into a host substrate. OSCAR mechanically encodes the ovipositor-inspired motion pattern through a spring-loaded cam system that drives twelve circumferential sliders in a phase-shifted reciprocating pattern, generating a controlled friction anisotropy at the interface that enables OSCAR to self-propel through the slippery, viscoelastic environment of the colon without requiring excessive normal loads. Experimental validation in \textit{ex-vivo} porcine colon demonstrated a steady-state traction force of $\approx 0.85~\mathrm{N}$ and an average speed of $3.08~\mathrm{mm/s}$, highlighting the potential of OSCAR to match standard safety and cecal intubation times of conventional colonoscopy.

Additionally, we presented an analytical framework that links cam kinematics to capsule thrust by capturing the viscoelastic stick--slip interaction at the tissue interface using a Kelvin--Voigt formulation, and validated this framework experimentally. The model revealed that thrust is largely speed-independent and scales linearly with cam duty-cycle asymmetry, underscoring OSCAR’s predictable performance and scalable design. Beyond OSCAR, the model provides a robust, device-agnostic tool to simulate and optimize future sliding-friction capsule designs.

By tuning friction anisotropy through mechanically-encoded motion patterns of sliders, OSCAR establishes a pathway toward friction-programmable robotic capsules that generate robust and predictable locomotion in the challenging environment of the colon and beyond, ultimately bringing painless robotic colonoscopy within reach.

\section{Acknowledgment}
We would like to thank the Delft University of Technology Central Workshop (DEMO) for their technical support in fabricating the prototypes.

\addtolength{\textheight}{-0cm} 

\bibliographystyle{ieeetr}
\bibliography{biblography}


\end{document}